\documentclass[letterpaper, 10 pt, conference]{ieeeconf}  

\IEEEoverridecommandlockouts                              
\usepackage{graphics} 
\usepackage{epsfig} 
\usepackage{times} 
\usepackage{amsmath} 
\usepackage{amssymb}  
\usepackage{bm}
\usepackage{bbm}
\usepackage{color, soul}
\usepackage{graphicx}
\usepackage[font=footnotesize,labelfont=bf]{caption}
\usepackage{subcaption}
\usepackage{cite}
\usepackage[table,xcdraw]{xcolor}
\usepackage{multirow}
\let\labelindent\relax
\usepackage{enumitem}
\usepackage{hyperref}

\usepackage{changes}
\definechangesauthor[name=Ioanna]{im}
\setdeletedmarkup{\color{red} \sout{#1}}
\setaddedmarkup{\color{cyan} #1}

\title{\LARGE \bf
Interpretability in Contact-Rich Manipulation via Kinodynamic Images }

\author{Ioanna Mitsioni$^{1}$, Joonatan Mänttäri$^{1}$, Yiannis Karayiannidis$^{2}$, John Folkesson$^{1}$ and Danica Kragic$^{1}$ 
\thanks{$^{1}$ Division of Robotics, Perception and Learning (RPL), CAS, EECS, KTH Royal Institute of Technology, Stockholm, Sweden 
        {\tt\small mitsioni,manttari,johnf,dani@kth.se}}%
\thanks{$^{2}$Division of  Systems and Control, Dept. of Electrical Engineering, Chalmers University of Technology, Gothenburg, Sweden 
        {\tt\small yiannis@chalmers.se}}%

}

\begin{document}
\maketitle
\thispagestyle{empty}

\begin{abstract}
Deep Neural Networks (NNs) have been widely utilized in contact-rich manipulation tasks to model the complicated contact dynamics. However, NN-based models are often difficult to decipher which can lead to seemingly inexplicable behaviors and unidentifiable failure cases. In this work, we address the interpretability of NN-based models by introducing the kinodynamic images.
We propose a methodology that creates images from the kinematic and dynamic data of a contact-rich manipulation task. Our formulation visually reflects the task's state by encoding its kinodynamic variations and temporal evolution. By using images as the state representation, we enable the application of interpretability modules that were previously limited to vision-based tasks. We use this representation to train Convolution-based Networks and we extract interpretations of the model's decisions with Grad-CAM, a technique that produces visual explanations.
Our method is versatile and can be applied to any classification problem using synchronous features in manipulation to visually interpret which parts of the input drive the model's decisions and distinguish its failure modes.
We evaluate this approach on two examples of real-world contact-rich manipulation: pushing and cutting, with known and unknown objects. Finally, we demonstrate that our method enables both detailed visual inspections of sequences in a task, as well as high-level evaluations of a model's behavior and tendencies. 
Data and code for this work are available at \cite{code}.
\end{abstract}

\section{Introduction}
Contact-rich tasks emerge naturally in many robotic applications, from pushing and grasping to in-hand manipulation. Modelling and describing the contact dynamics analytically is especially challenging due to their frictional nature, discontinuities and variety of surfaces and objects. Moreover, the dynamics can vary throughout the task execution on the same object. In cutting for example, changes in the knife contact surface area, the width of the inserted blade, and transient physical properties of the cutting object will result in temporally and spatially varying dynamics.

Models based on neural networks can handle the heterogeneous kinematic and dynamic data of a manipulation task and discover relevant patterns. 
They can account for variations in the objects and modes of interaction, generalize to unseen objects and address the diverse dynamics  \cite{bohgGraspSurvey, slipDet, PomdpHapticData, dressing, Nagabandi2019, openAIinhand}. 
However, small differences in training parameters or network architectures can result in significantly different behaviors. Additionally, due to the deep networks' black-box nature, the increased modelling performance comes at the cost of interpretability. For human operators it is mostly unclear why an input leads to a specific output, which is crucial both for safety and performance when the models are deployed within a real robotic system. 

\begin{figure}[t]
\centering 
\begin{tabular}{cc}
     \includegraphics[width = 0.4\linewidth]{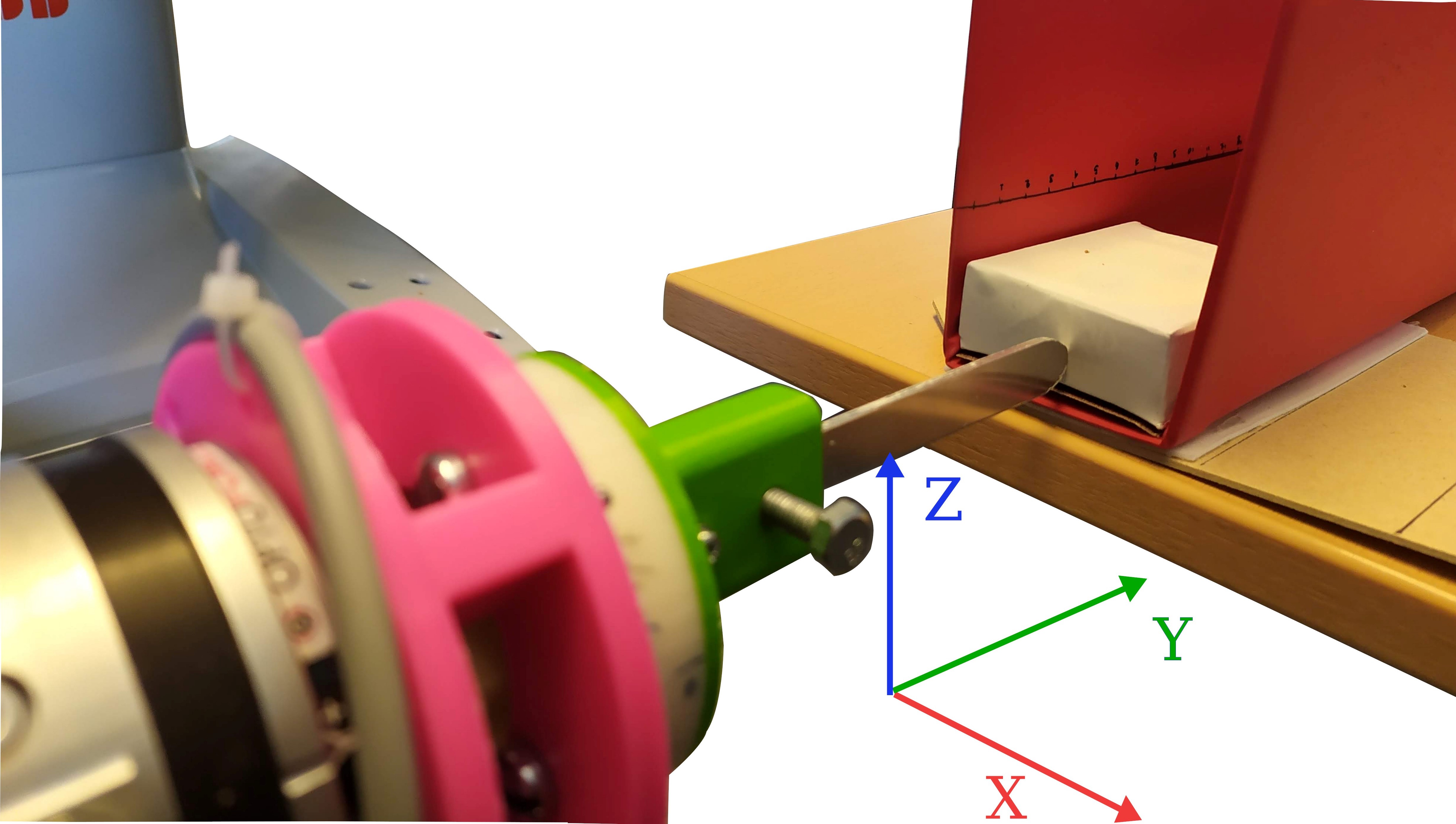}&  
     \includegraphics[width = 0.4\linewidth]{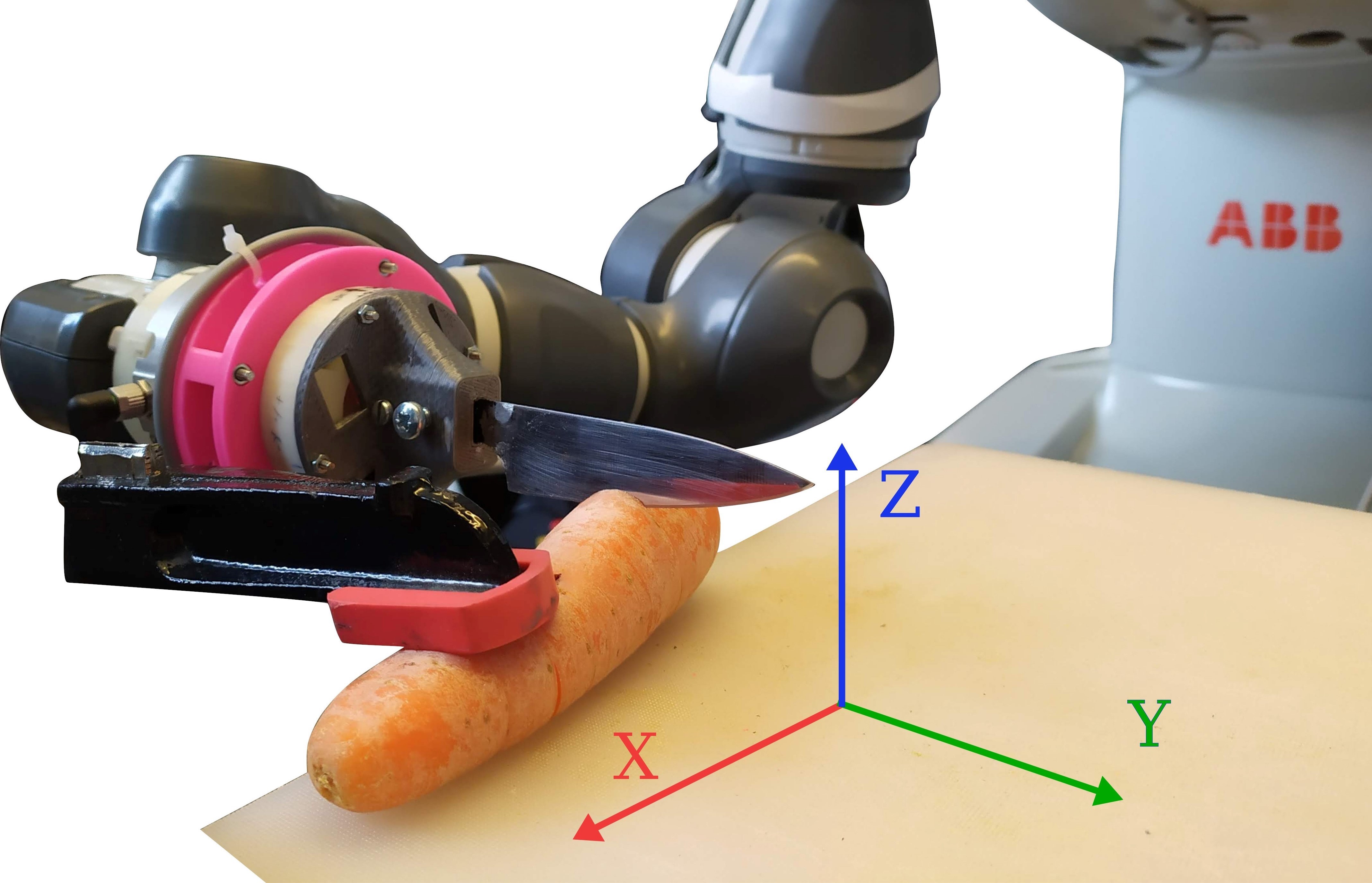}\\ 
     \footnotesize (a) Pushing task & \footnotesize (b) Cutting task \\
\end{tabular}
\begin{tabular}{c}
     \includegraphics[width = 0.9\linewidth]{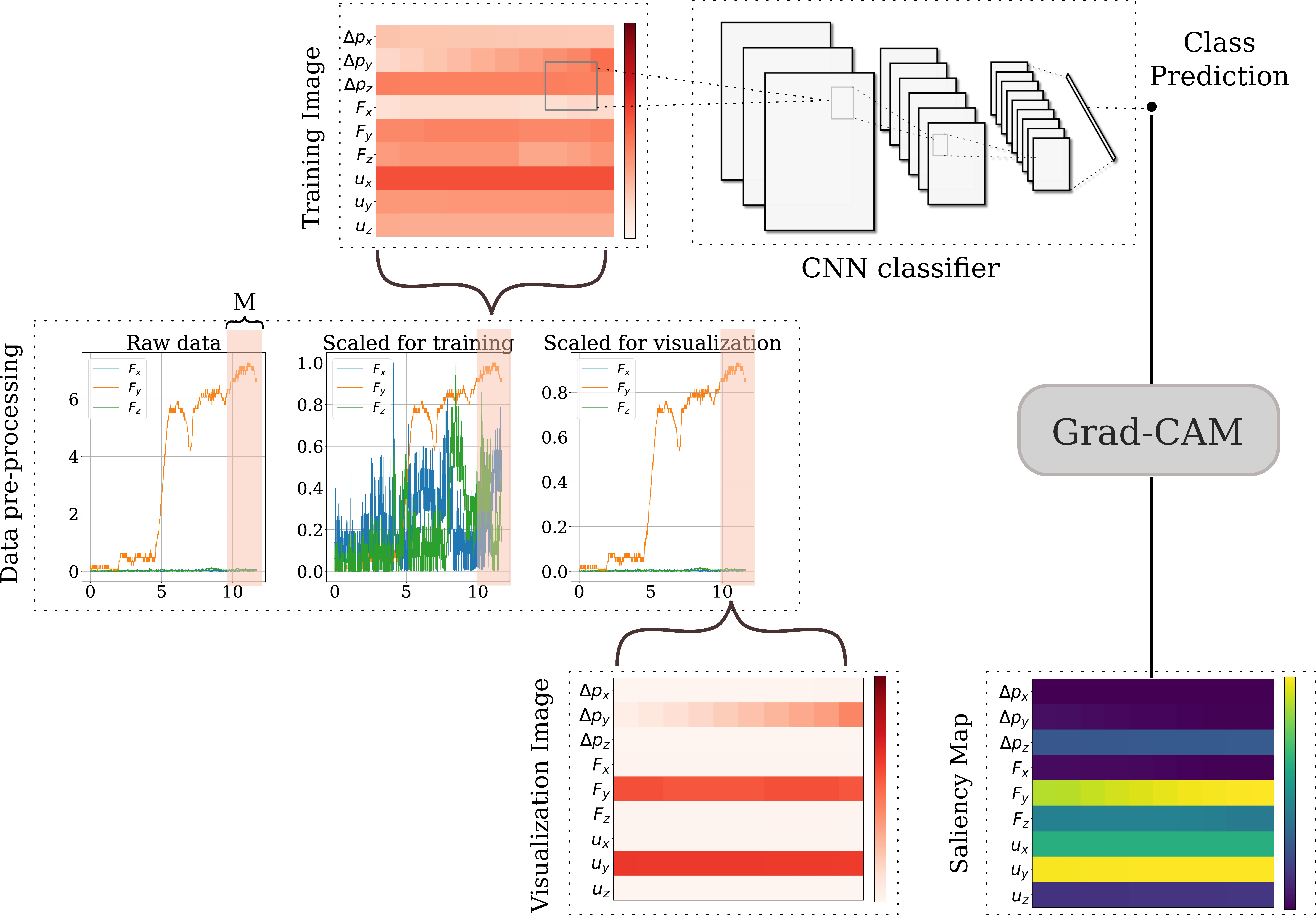} 
      \\ \footnotesize (c) System overview 
\end{tabular}
\caption{\footnotesize To create the interpretable kinodynamic image, we transform state and input sequences of length $M$ into a pixelated image where the placement of the pixels is determined by the feature and the timestep, while the color by the magnitude. Every kinodynamic image corresponds to a block of measurements that starts at the current timestep $t$ and ends at $t+M$. After scaling the images for training and visualization accordingly, we use them as inputs to a CNN that predicts the class of a future sequence. Lastly, we extract saliency maps through Grad-CAM that answer the question \textit{"Which features led the network to this prediction?"}.}
\label{fig:overview}
\vspace{-7mm}
\end{figure}

Consider the example of food-cutting tasks with objects of various shapes and stiffness. During execution, different motions might result in successful cuts or failed ones where the knife gets stuck in the object and potentially leads to a mechanical shutdown due to excessive forces.
To describe this interaction, we need to take into account the robot's kinematics and the dynamics depending on the knife's shape, the materials and the controller parameters. Deriving
an analytical model to determine the motion's outcome and encompass all the possible variations is practically impossible. 

\begin{figure*}[h]
\centering
\begin{tabular}{cccc}
\includegraphics[width=0.25\textwidth]{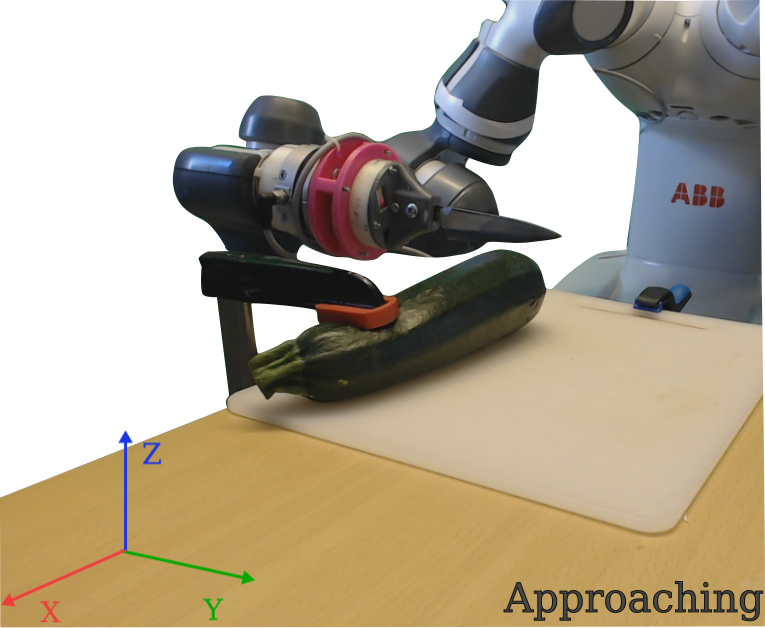} &
\hspace{-4.7mm}
\includegraphics[width=0.25\textwidth]{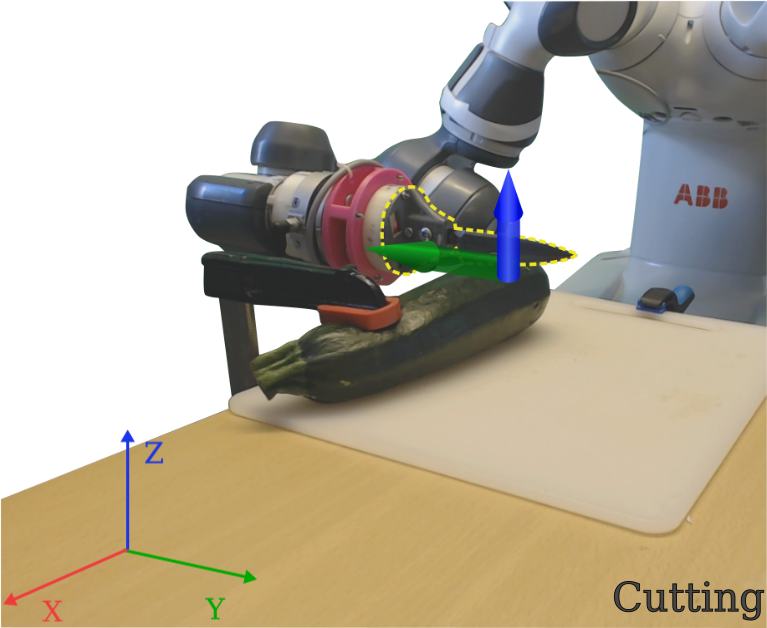} &
\hspace{-4.7mm}
\includegraphics[width=0.25\textwidth]{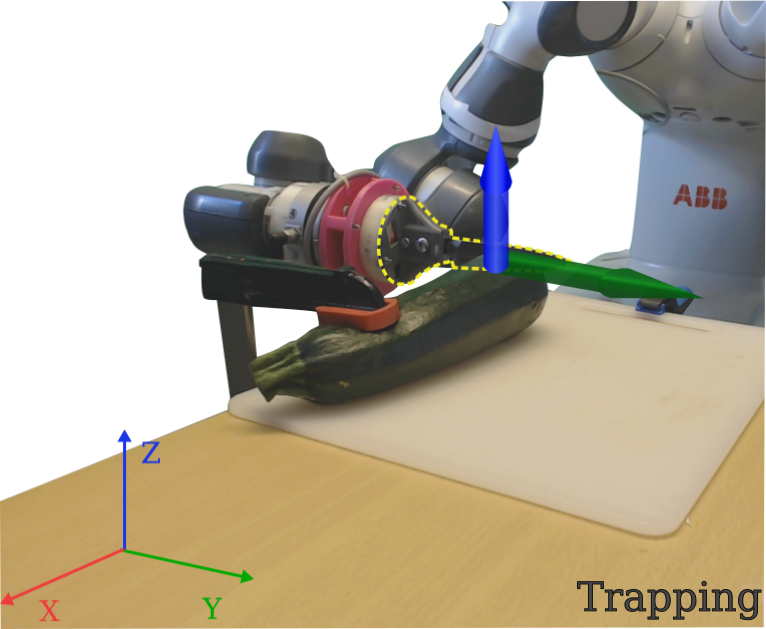} \\
\end{tabular}
\begin{tabular}{cccc}
\includegraphics[width=0.25\textwidth]{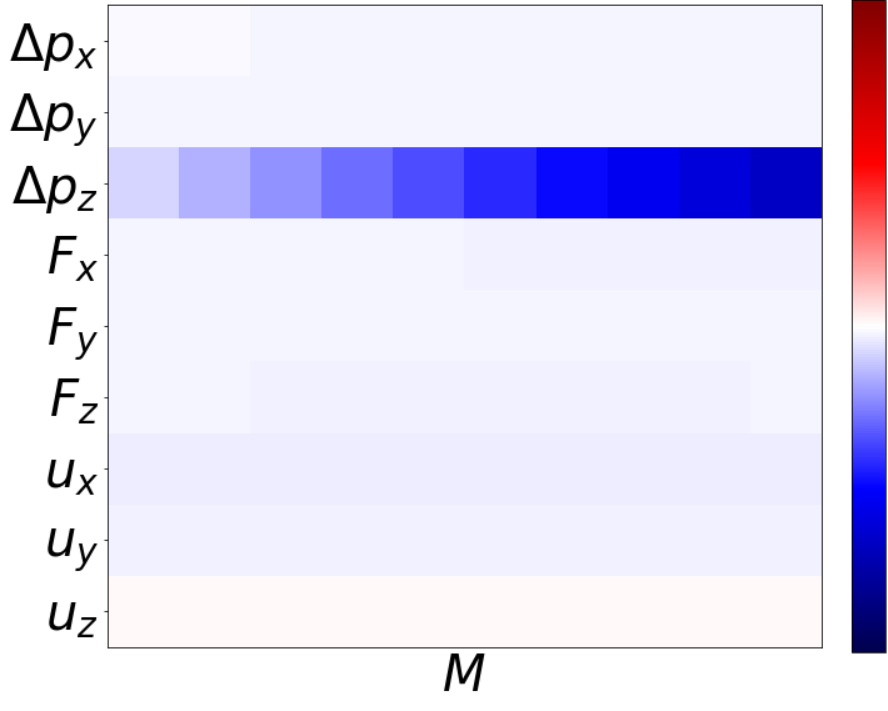} &
\hspace{-4.7mm}
\includegraphics[width=0.25\textwidth]{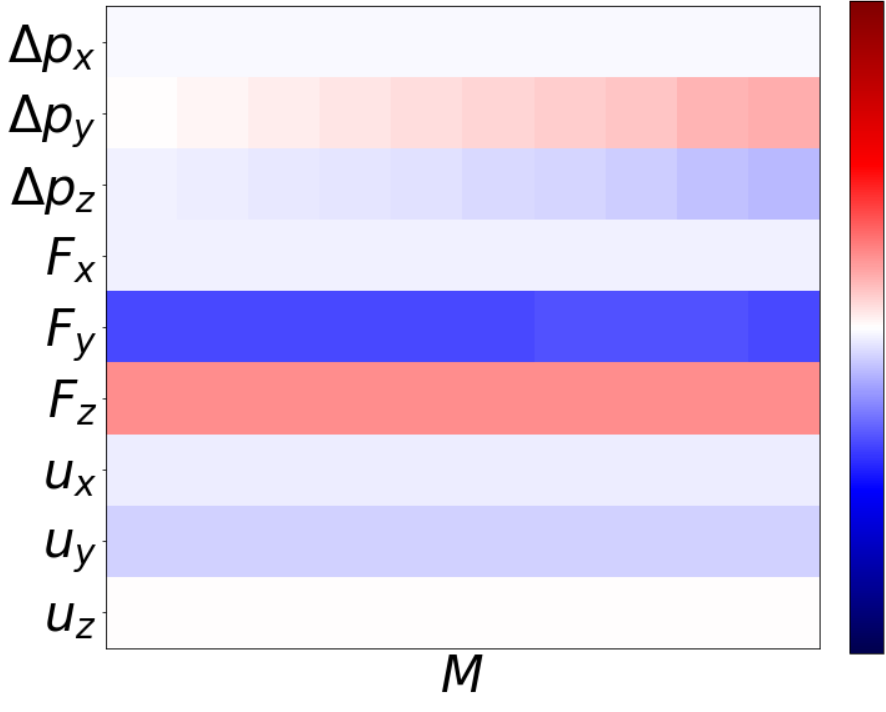} &
\hspace{-4.7mm}
\includegraphics[width=0.25\textwidth]{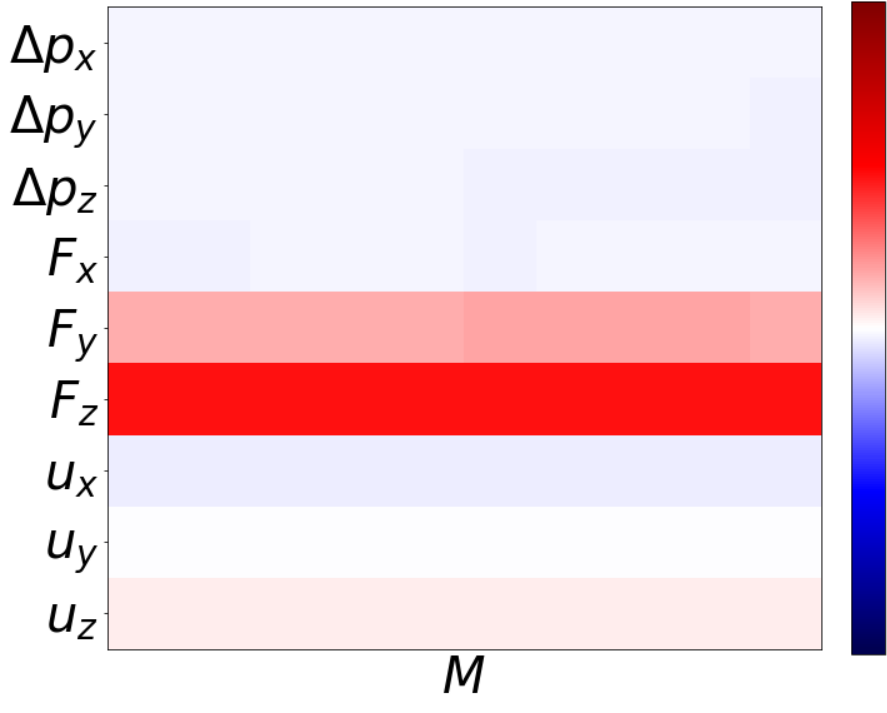} \\
\textbf{(a)}  & \textbf{(b)} & \textbf{(c)}  \\[6pt]
\end{tabular}
\caption{ \footnotesize Visualization of a failed cutting trial. The kinodynamic images correspond to the sequence (block) of measurements at the specific time instance: \textbf{(a)} While the robot is approaching the object, there is free motion on axis Z ($\Delta p_z$) and no contact forces.
\textbf{(b)} When the robot is able to cut through the object, there is sawing motion ($\Delta p_y$), slicing motion ($\Delta p_z$), as well as forces on both axes, $F_y, F_z$.
\textbf{(c)} When the controller gains are not appropriate for the contact's stiffness and friction, the knife is trapped inside the object, forces $F_y, F_z$ are high and a mechanical shutdown might occur.  
The colormaps represent the magnitude for every feature on a normalized scale spanning from blue for the largest negative to red for the largest positive value.}

\label{fig:Stuckage}
\vspace{-5mm}
\end{figure*}

Nevertheless, we can train a network to timely predict an undesirable event to allow for corrective actions through a planning module or through different controller parameters. On one hand, these system reactions are helpful, but should be triggered only when necessary since otherwise they inhibit normal task execution. On the other hand, if not triggered in time, they will not prevent failure. In both cases, it is imperative to understand \textit{why} they are either needlessly or untimely triggered and address the failure modes. 

The ability to explain how the system's inputs trigger specific answers enhances the utility of data-driven models. Employing interpretability modules can reveal that a model is sensitive to a sensor's noise and will be biased by it during the task execution. Interpretability can also highlight that two models that seem identical are utilizing different features to make decisions or why a model cannot generalize to new situations. Concretely, it offers a deeper understanding of how the data-driven models operate and allows the user to thoroughly evaluate and improve them according to the task.  

To explain how an answer was produced, it is necessary to inspect the network's activations with respect to their predictions. This inspection however is intractable for most models if the activations and the inputs are not in an image format. To address this, we transcribe the problem into an image domain to allow its visualization through the method summarized in Fig. \ref{fig:overview}.
We propose a way of constructing the images so that they are visually intuitive, compact, easily modifiable to include more features and allow us to trace the activations back to the input with no special operations that could limit their applicability.
The generated images of kinematic and dynamic data, called kinodynamic images, depict the state of the system during a sequence of timesteps and permit individual visual inspection of the features as shown in Fig. \ref{fig:Stuckage}. Furthermore, we propose to use the images with a CNN architecture that is not constrained by the placement of the features in the image.
To examine which parts of the input contribute to the network's answers, we employ Grad-CAM~\cite{GradCam}, a technique for producing visual explanations in the form of saliency maps.

In summary, we present a method for interpreting learned models in manipulation. Our method can be applied to any manipulation task that is formulated as a classification problem. Moreover, it can utilize a variety of features stemming from proprioception or kinesthetic sources, as long as they are synchronous.
We demonstrate how to construct  kinodynamic images and consequently train and evaluate models for classification in two examples of contact-rich manipulation; pushing and cutting. For both tasks, the class to predict denotes the continuation of motion in a future instance. In our experiments, we show examples of different behaviors from seemingly identical models and how to detect them through visual interpretation of isolated sequences or comparisons of the overall feature utilisation. 

The main contributions of this work are listed below:
\begin{enumerate}[label=(\alph*)]
    \item It is the first work that treats the interpretability of neural networks in contact-rich manipulation tasks.
    \item It produces qualitative, visual interpretations of isolated sequences in a task's evolution, such as transitions between states, for thorough inspection.
    \item It can also be used to produce high-level, statistical results for a model's behavior, such as overall feature utilization, on the entirety of collected data.
\end{enumerate}

\section{Related Work}\label{relatedwork}
\paragraph*{Vision techniques in contact-rich manipulation}
Image-like structures have been explored in contact-rich tasks to leverage the modelling capabilities of CNNs, usually in the form of tactile data, allowing vision techniques to be applied to touch. 
In \cite{learningTactile1}, the spatial signals from a BioTac sensor are used to estimate contact forces while in \cite{learningTactile2}, the authors employ an optical tactile sensor to train a CNN for edge perception and contour following. A more general approach is presented in \cite{gelsight} where the observations from a GelSight sensor are used to plan explicitly in the raw tactile observation space and manipulate an object through touch. A multimodal representation for contact-rich tasks has been proposed in \cite{makingsense} to encode heterogeneous sensory input that includes forces, end-effector positions and velocities, as well as visual data, to learn policies for a peg insertion task. While all of these works address contact-rich tasks, they lack explanations about the underlying decision-making of the networks and only utilize them to obtain generalizable solutions for manipulation.
Lastly, the work in \cite{DeepInsight} directly addresses the transformation of non-image data to an image to leverage a CNN architecture in several non-robotic tasks. This work is not focusing on interpretability either and for the image construction, a dimensionality reduction technique is first used to encode the data into a 2-dimensional format.
As opposed to reducing the data dimensions, we construct the kinodynamic images to preserve the original dimensionality, which is important for visual interpretation in the Cartesian task space. When encoding data to a latent space that is produced by linear combinations of all the input features, the representation of the spatial and color dimensions of the corresponding pixels lacks physical meaning w.r.t. the interaction controller quantities. Even if the operation is fully reversible and we can return to the original raw data format, it is not straightforward to track which units have contributed to the network's decision, thus losing the ability to readily extract an explanation in the world frame.

\paragraph*{Interpretability methods} 
Interpretability of deep neural networks has recently become a focus point in the deep learning community. The furthest progress has been made for
classification networks operating on single images \cite{Simonyan2014DeepMaps, Zeiler2014, Montavon2018, tcavglobal_icml18}. Interpretability methods can be divided into two categories: network-centric and data-centric. Network-centric methods focus on specific units in the model, for example Activity Maximization \cite{Erhan2009VisualizingNetwork}. This method aims to find what types of inputs would maximize the activation of a specific unit by using gradient ascent w.r.t. the general input space, not necessarily the current input. In this sense, it is more appropriate for examining network architectures and discerning the optimal parts in the entirety of the input space, as opposed to inspecting the actual inputs for a set architecture, which is our goal.
Data-centric methods on the other hand, focus on examining or manipulating input data to determine which patterns the model has learned for the task. Some examples of these methods are Layer-wise Relevance Propagation \cite{Montavon2018}, Excitation Backprop \cite{ExcitationBackprop16}, and Grad-CAM \cite{GradCam}. The first two utilize activations and weights which are normalized and backpropagated, while Grad-CAM calculates saliency maps using activations and gradients. A side-effect of how the first two methods operate is that they produce more fine-grained answers, in contrast to Grad-CAM which is more appropriate for detecting regions of interest in the input space. Because we are interested in considering the whole sequence of the feature (a row in the image), this is beneficial for the images we construct.
In addition, in a work evaluating the scope and quality of explanation methods \cite{adebayoneurips18}, Grad-CAM was found to be one of the few methods that take into account both the input data and the model parameters and do not operate similarly to an edge detector.

\paragraph*{Interpretability in robotic tasks}
Interpretability in robotics has mainly been explored in human-robot interaction as a tool to align human intuition with what the robot learns. A method utilising user-defined symbols during learning from demonstration to produce a semantically aligned latent space has been proposed in \cite{hristov2019disentangled}, while \cite{kulkarni2020designing}, a way of designing the environment of the robot to allow for long-term interpretable plans is introduced.
In another example of human-robot collaboration \cite{interpretableCollab}, the authors consider interpretability in the context of human motion recognition. They employ skeletal and object position data and introduce a human activity classifier that allows queries over anomalies in the input signal and can provide feedback in a human-interpretable manner.
The concept of interpretability has also been examined in \cite{dotToDot} as part of a hierarchical reinforcement learning structure with two agents. The high-level agent is deconstructing the current problem into sub-goals for the low-level agent to reach. By using this hierarchy, they are able to visualize the parts of the environment that are considered prime candidates for the sub-goal and evaluate their intuitiveness.
In a block stacking setup, \cite{prospection} presented an inference method for interpretable plans from natural language, proprioceptive and image data. In \cite{Ehsani_2020_CVPR}, they authors explained an object motion in human-object interaction by training networks to predict contact forces and contact points.
Lastly, \cite{wormdude} proposed a network design methodology for end-to-end robotic control that produces compact networks with interpretable structure, thanks to their dedicated structures. 
In contrast, we focus on low-level interpretations to understand why and how a trained model works within a manipulation task.



\section{Method}\label{Section:KinodynamicImage}
We assume a robotic manipulator equipped with external force sensing and that the task evolution depends on the state $\mathbf{x}$ and the control action $\mathbf{u}$. We further assume that the learning objective can be formulated as a classification problem of the form: \textit{given the current state and control input, what is the class of the future state?}. To model the state $\mathbf{x}$ of the interaction task we use the displacement of the end-effector and the sensed forces as they can encapsulate the variations in the friction and stiffness of the contact. The control input for our set of experiments was measured from an admittance controller presented in \cite{mitsioni2020modelling}. To produce explanations for a prediction, sequences of raw data are first transformed into images which are scaled for training and used as inputs for a CNN. A saliency map of the feature importance is then produced by Grad-CAM. To facilitate the interpretation, the saliency maps are presented together with images that have been scaled to be visually intuitive. Detailed descriptions of the individual components are presented in the following subsections.

\paragraph*{Creating the Kinodynamic Image}
To create the network's input, we form blocks $b$ of the tuple $(\mathbf{x}, \mathbf{u})$ that consist of non-overlapping sequences of measurements of length $M \in \mathbb{N}$. We denote the sequence of end-effector positions during a block $b$ as $\mathbf{p}_b$, the sequence of sensed forces as $\mathbf{F}_b$ and the control input as $\mathbf{u}_b$. We then define the state as a combination of the relative change in positions $\Delta\mathbf{p}_b \in \mathbb{R}^{3\times M}$ between blocks and the sensor's forces $\mathbf{{F}}_{b} \in \mathbb{R}^{3\times M}$ such that $\mathbf{x} = [\Delta\mathbf{p}_b, \mathbf{{F}}_{b}]^T \in \mathbb{R}^{M\times 6}$, while the control input $\mathbf{u} = \mathbf{u}_b^T  \in \mathbb{R}^{M\times 3}$. 

Every input to the network thus consists of $M \times L$ measurements with $L = 9$. To form the kinodynamic images, we transform these inputs into an image with width $M$ for the temporal evolution, and height $L$ for the total amount of features. Consequently, we create an $L \times M$ grid of pixels whose color is given by encoding the magnitude of the feature during that timestep into a 3-channel RGB value via a colormap. This procedure does not depend on the specific features or controller used and can be applied to a multitude of tasks. In addition, it can be used for inputs of higher dimensionality, such as the joint states of a dual-arm manipulator, simply by adjusting the height $L$ of the image.

In the data preprocessing step, we utilize two different scalers; one for training and one for visualization.
\begin{figure}
	\centering
    	\subfloat[\footnotesize Raw data, training scaler, visualization scaler on the original signal for a pushing task. ]{\includegraphics[width=0.95\linewidth]{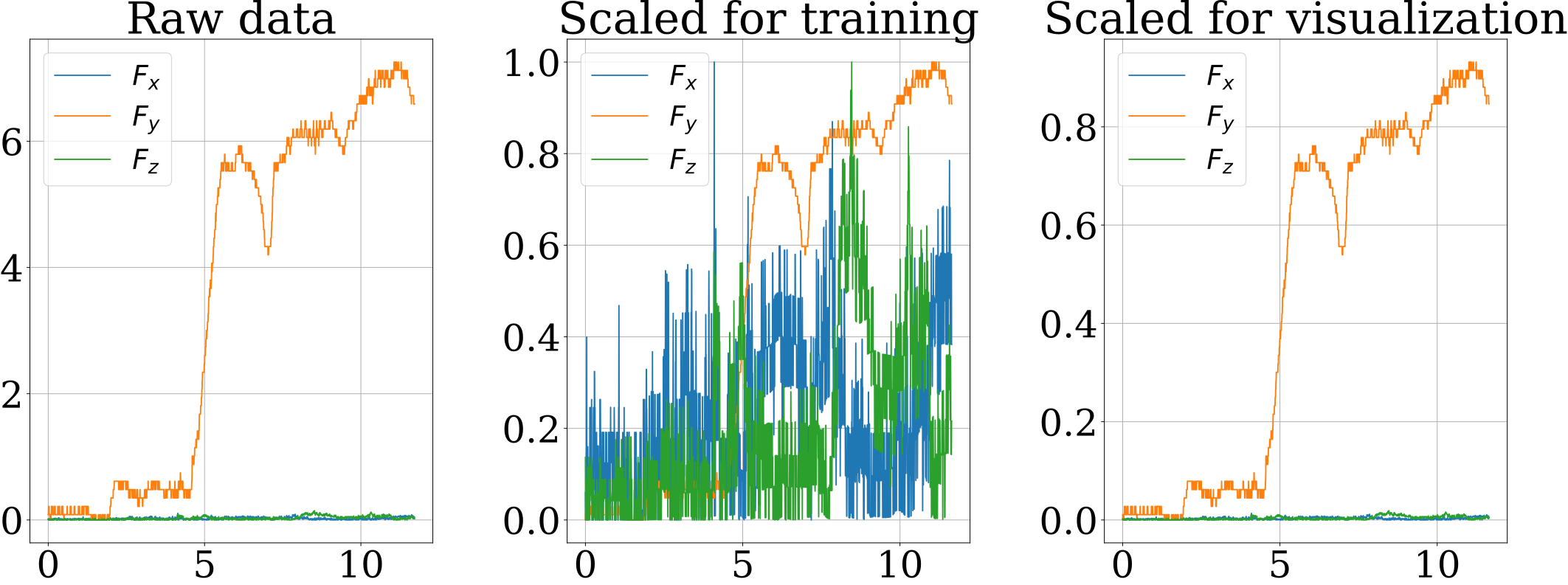}\label{fig:raw_signal}}
    	\\
    	\subfloat[\footnotesize A block of measurements taken from the above signal scaled with the training scaler (left) and  visualization scaler (right).]{\includegraphics[width=0.95\linewidth]{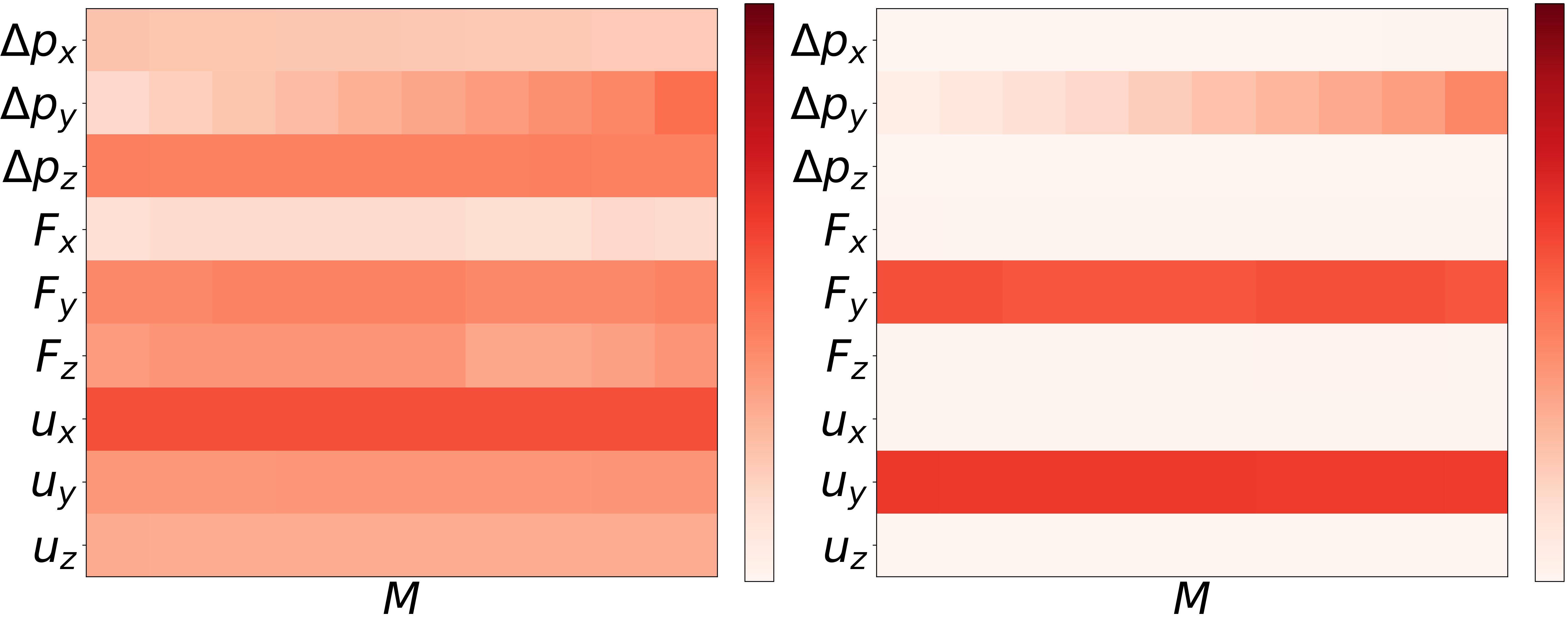}\label{fig:scale_train}} 
	\caption{\footnotesize Scaler differences in signal form (top) and image form (bottom).}
	\label{fig:scalercomparison} 
	\vspace{-5mm}
\end{figure}
The first one is a standard scaler that scales every feature separately across the training set to ensure equal consideration by the network.
The standard scaling operation has the disadvantage of producing images that are not visually intuitive as every component of the features will now be in the same value range. This implies that the noise in irrelevant to the task axes, will now visually appear significant. An example of this can be seen in Fig. \ref{fig:scalercomparison}. The actual force signals resulting from motion on axis $Y$ are shown in the leftmost plot of Fig. \ref{fig:raw_signal}. 
In the middle plot, the signals have been scaled for training and whatever noise resided on axes $X$ and $Z$ is now independently scaled to match the feature range, which produces visually uninformative images such as the left one in Fig. \ref{fig:scale_train}.

The second scaler is strictly used for visualization purposes and scales the features of every group according to the absolute maximum value observed for the group across the training data. Concretely, during training the features are scaled row-wise (with respect to the image), while for visualization, we scale the groups of features that correspond to displacements, forces and inputs based on the largest value measured within the same group.

In both cases the data are scaled to the the interval $(0,1)$, since we employ heterogeneous features with different ranges. The last figure in Fig. \ref{fig:raw_signal} depicts the same force signals but scaled to preserve their relative magnitude, resulting in the more informative and intuitive kinodynamic image on the right in Fig. \ref{fig:scale_train}.

\paragraph*{Classification Network}
We used two variants of CNN networks, a purely Convolutional model, and a Convolutional Long-Short Term Memory (C-LSTM) model. The convolutional architecture consists of three 1D convolutional layers followed by a fully-connected layer and a Leaky ReLU after the second layer. The convolutional layers have 32, 64 and 128 output channels, with filter sizes (1,5), (1,3), (1,1) of single stride and no padding. The C-LSTM architecture has 3 C-LSTM layers, each with 32 channels and filter size (1,5) and (0,2) padding, operating on a sequence of the 3 most recent images. The filter sizes and their strides were chosen to reflect the shape of the individual features in the image, produce crisp salient maps and avoid constraints on the feature placement. Consecutive features can still contribute simultaneously to the prediction, but they produce individual salient regions.
For training, we used a 75-25 train-test split and the Cross-Entropy loss.
During every pass, the convolutional layers treat time sub-sequences of individual features and their output is then used to produce saliency maps centered at those features. 
This is a result of Grad-CAM using the gradient of the target class logit in the fully-connected layer w.r.t. the activations in the final convolutional layer to generate a heat map that highlights the important regions in the image for a prediction. 
In essence, these saliency maps refer to the future block and show the parts of the current one that contributed the most to the prediction.

\begin{figure*}
	\centering
    	\subfloat[$b-1$]{\includegraphics[width=0.33\textwidth]{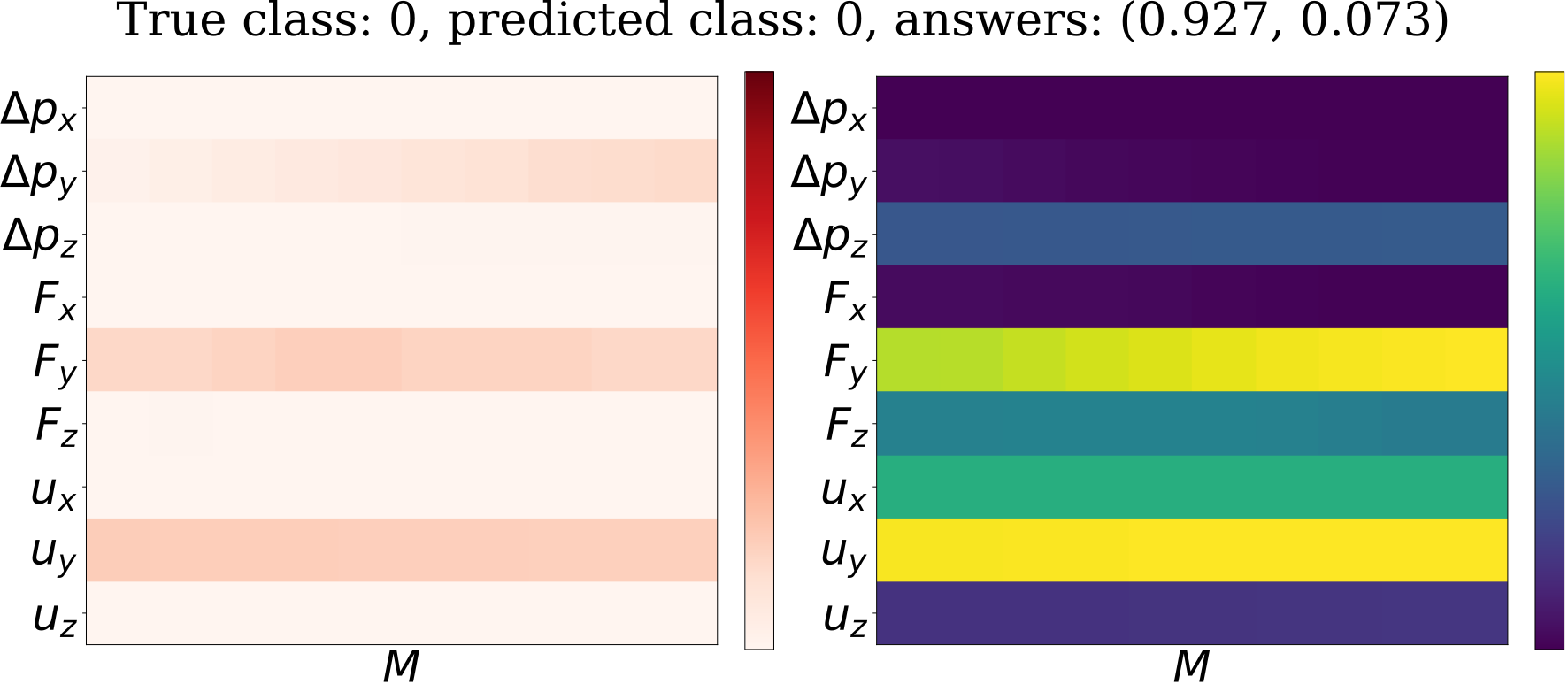}}
    	\subfloat[$b$]{\includegraphics[width=0.33\textwidth]{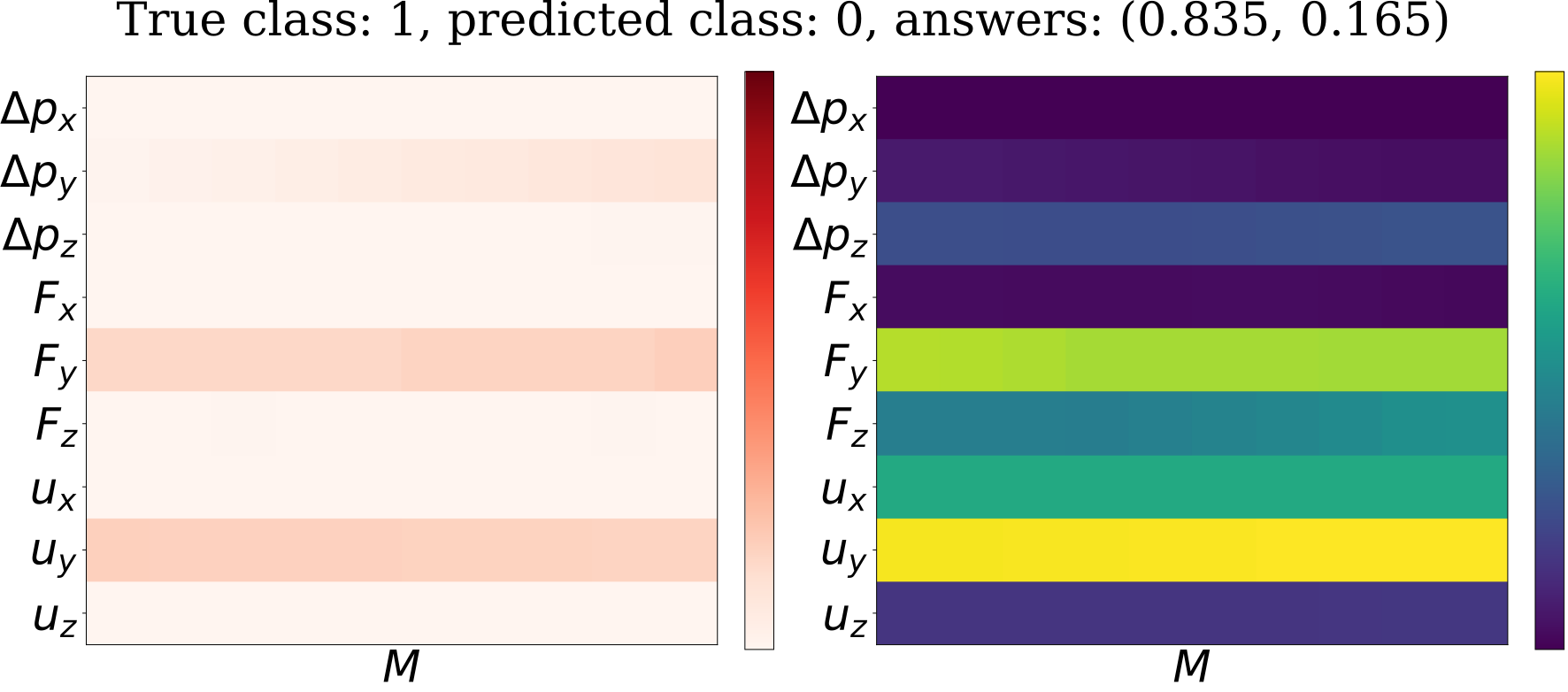}}
    	\subfloat[$b+4$]{\includegraphics[width=0.33\textwidth]{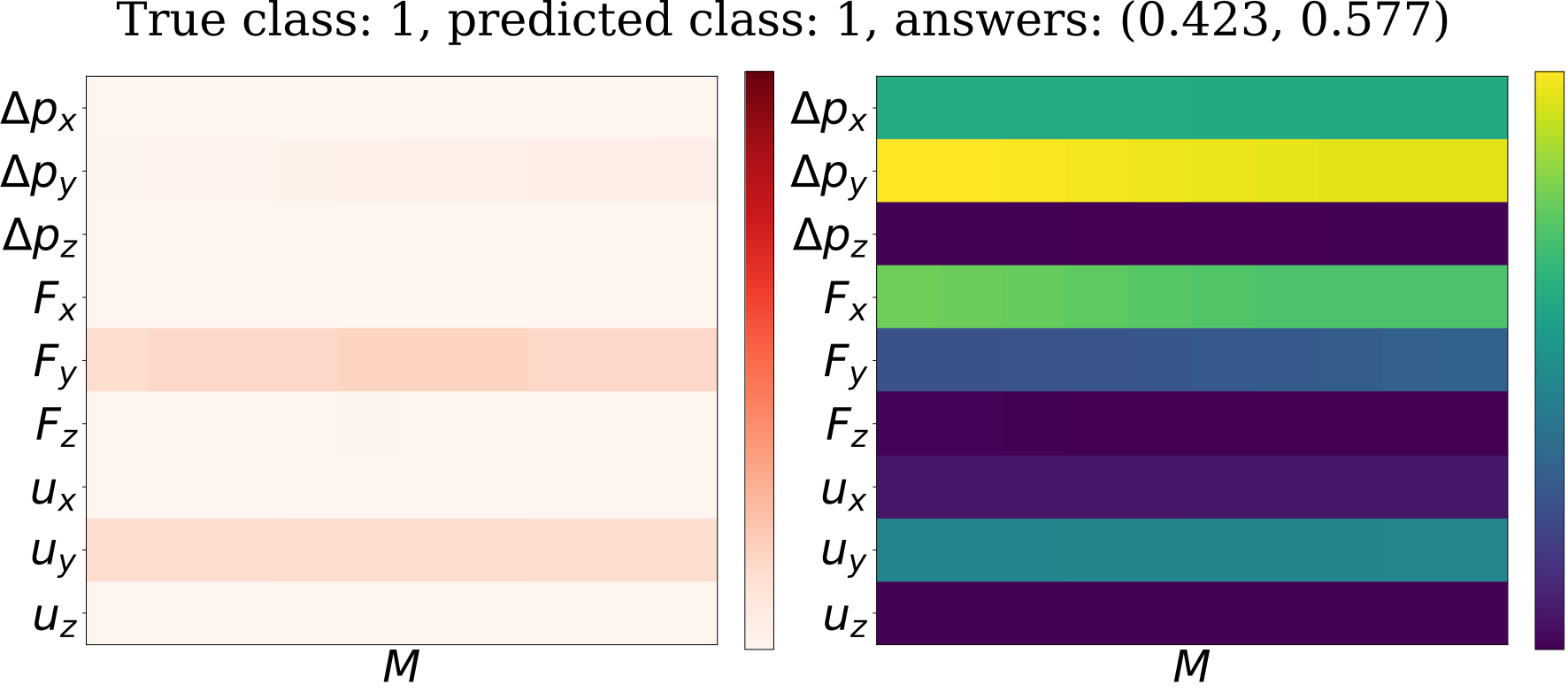}}
	\caption{\footnotesize A sequence of input frames at blocks $b\pm i$ during a transition for the pushing task. The class prediction shifts from True Positive, through False Positive until it reaches True Negative.}
	\label{fig:1Dcrit} 
	\vspace{-5mm}
\end{figure*}


\section{Evaluation}
\paragraph*{Dataset Collection}
To collect data, we employ the same controller as in \cite{mitsioni2020modelling}. 
Specifically, if we denote the desired positions and velocities of a trajectory as $\mathbf{p}_d, \dot{\mathbf{p}}_d \in \mathbb{R}^3$, the reference and measured forces as $\mathbf{f}_s, \mathbf{f}_r$ and the goal is to track the trajectory compliantly, we can define the velocity control input as $\mathbf{u} = \mathbf{K}_a(\mathbf{f}_s - \mathbf{f}_r)$.
We can then define the desired compliant behavior as $\mathbf{f}_r =  \mathbf{K}_a^{-1}(\mathbf{K}_p \mathbf{e} -  \dot{\mathbf{p}}_d)$, where $\mathbf{K}_p,\, \mathbf{K}_a \in \mathbb{R}^{3\times 3} $ are the stiffness and compliance gain matrices and $\mathbf{e} = \mathbf{p}-\mathbf{p}_d$ the position error. The controller operates at $200 \textrm{Hz}$, so a block of length $M$ corresponds to a duration of $M/200 \textrm{s}$.

\begin{table}
    \subfloat[\footnotesize Training parameters.]{
	\resizebox{0.95\linewidth}{!}{%
    \begin{tabular}{c c c c c}
    \hline
    \textbf{Task} & \textbf{Network} & \textbf{Learning rate} & \textbf{Epochs} & $\mathbf{H_{block}}$ \\ \hline
    Push & \texttt{EventNetP} & 3e-04 & 30 & 5 \\ 
    Cut & \texttt{EventNetC1} & 1e-03 & 25 & 3 \\ 
    Cut & \texttt{EventNetC2} &  5e-04 & 25 & 3 \\
    Cut & \texttt{EvenNetCLSTM} & 1e-03 & 25 & 3 \\
    Cut & \texttt{EventNetGen} & 1e-03 & 25 & 2 \\
    \hline
    \end{tabular}%
    }\label{tab:training_params}}
    \vspace{2pt}
    \\
    \subfloat[\footnotesize Training and testing scores.]{
    \resizebox{0.95\linewidth}{!}{%
\begin{tabular}{lcc|cc}
\hline
& \multicolumn{2}{c|}{ $\mathbf{F_1}$\textbf{-score} (Train)}  & \multicolumn{2}{c}{ $\mathbf{F_1}$\textbf{-score} (Test)} \\
& C = 0                       & C = 1   & C = 0                    & C = 1     \\ \hline
\multicolumn{1}{l|}{\texttt{EventNetP}}     & 0.99                        & 0.95                       & 0.97                     & 0.95                    \\
\multicolumn{1}{l|}{\texttt{EventNetC1}}    & 0.85                        & 0.82                       & 0.85                     & 0.85                    \\
\multicolumn{1}{l|}{\texttt{EventNetC2}}    & 0.85                        & 0.82                       & 0.84                     & 0.85                    \\
\multicolumn{1}{l|}{\texttt{EventNetCLSTM}} & 0.87                       &          0.82                  &         0.83                 &          0.84               \\
\multicolumn{1}{l|}{\texttt{EventNetGen}}    & 0.85                        & 0.82                       & 0.91                     & 0.55                    \\ \hline
\end{tabular}%
    }\label{tab:training_results}}
    \caption{\footnotesize Training and performance details.}
    \vspace{-6mm}
\end{table}

For the pushing task, we keep the compliance at $\mathbf{K}_a = 0.005$ and the desired trajectory consists of three pushing phases that start and end with zero velocity for a total of $20 \textrm{s}$.
The pushed box weighs $190 \textrm{g}$ and to create variation in the dynamics, we add weights to the box ($275, 370, 645 \textrm{g}$) and use five different materials on the surface of the motion. 
The surfaces used were cork, sandpaper, felt, gouache paper and smooth crafting paper for a total of 48 datasets. To ensure the object will not rotate during the execution or tip upwards, all the experiments are performed inside a track that constrains the motion on the other axes and the pushing point is as close to the center of the object as possible. Datasets that include accidental rotations or significant frictional forces on the other axes were discarded. 
For the cutting task, the trajectories consist of periodic sawing motions on axis $Y$, while moving downwards on axis $Z$. The datasets collected for this task comprise 43 cutting trials on 3 different objects: zucchinis, potatoes and carrots. To create variation in the dynamics, we vary the stiffness gain of the controller $\mathbf{K}_a$ in the range $[0.0001-0.01]$ as well the commanded trajectories to reflect a range of more to less aggressive cutting strategies by changing the duration of each motion to be either $2, 3$ or $4 \textrm{s}$, the sawing range $(3, 4, 5 \textrm{cm})$, the number of sawing-slicing repetitions, ($1,2,3$), and the slicing distance in every repetition $(1, 2 \textrm{cm})$. 

\begin{figure*}
    \centering
    	\subfloat[$b-4$]{\includegraphics[width=0.25\textwidth]{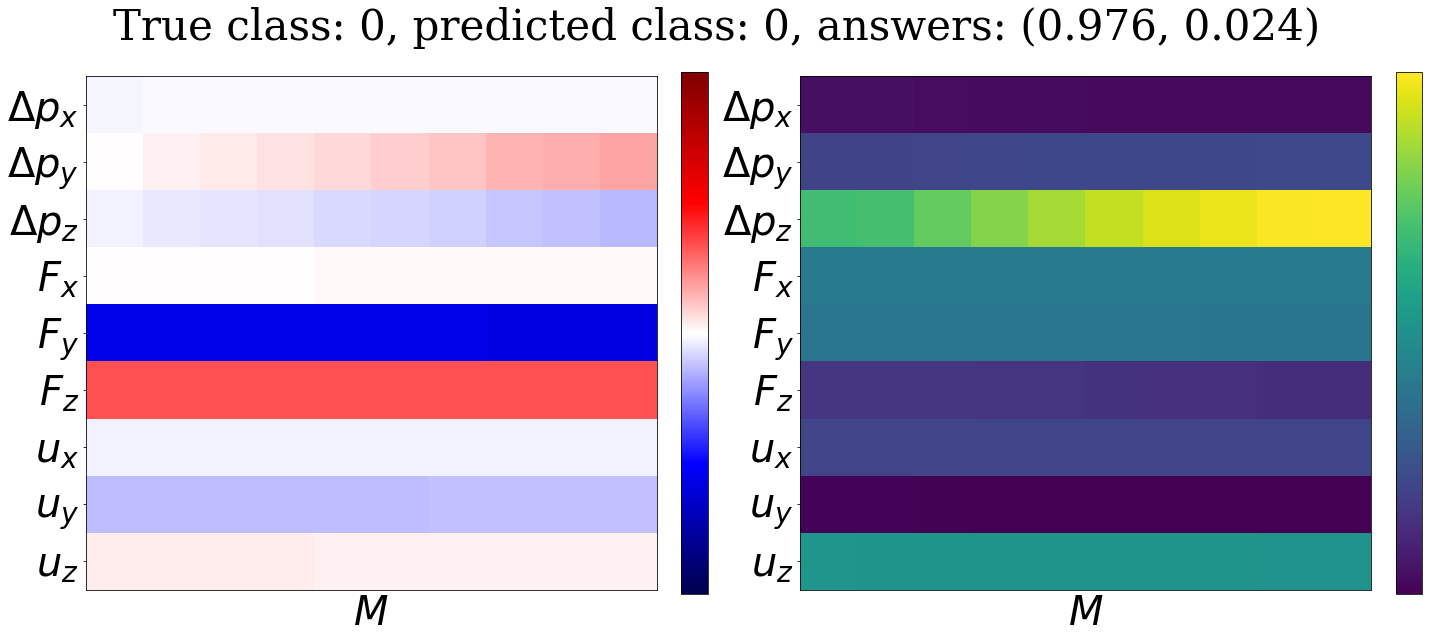}\label{fig:EventNetC1_a}}
    	\subfloat[$b-1$]{\includegraphics[width=0.25\textwidth]{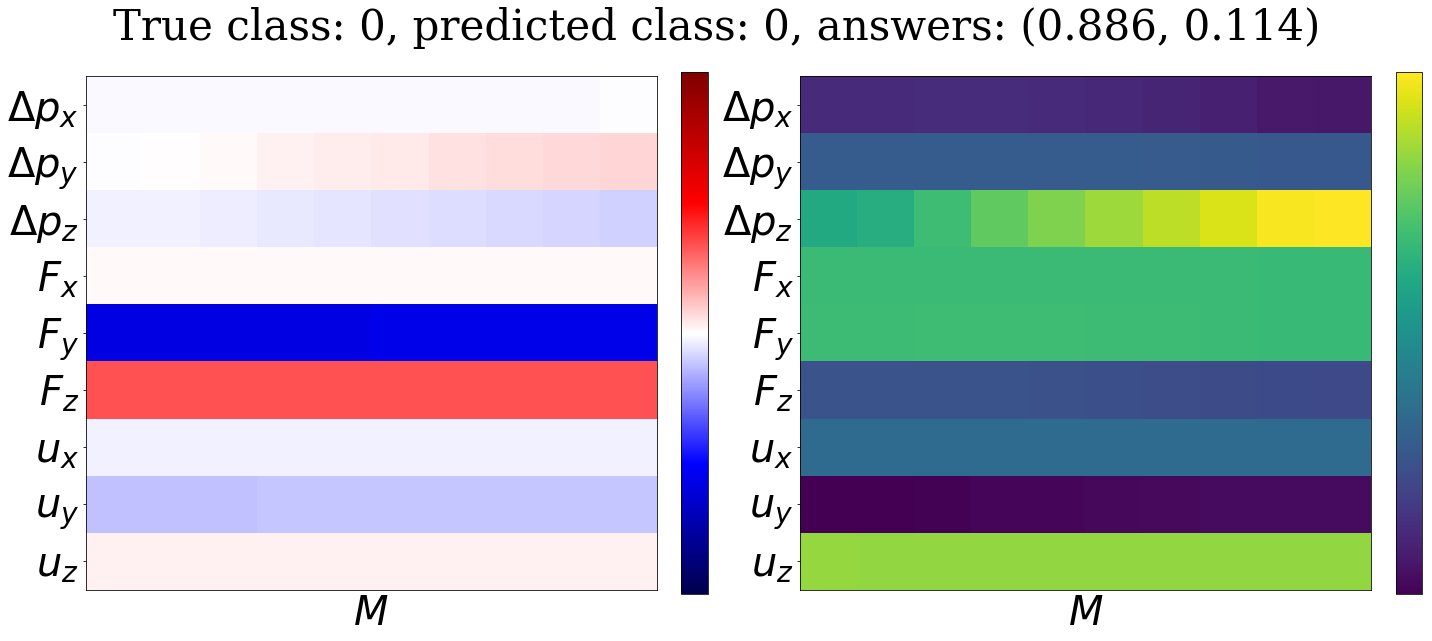}\label{fig:EventNetC1_b}}
    	\subfloat[$b$]{\includegraphics[width=0.25\textwidth]{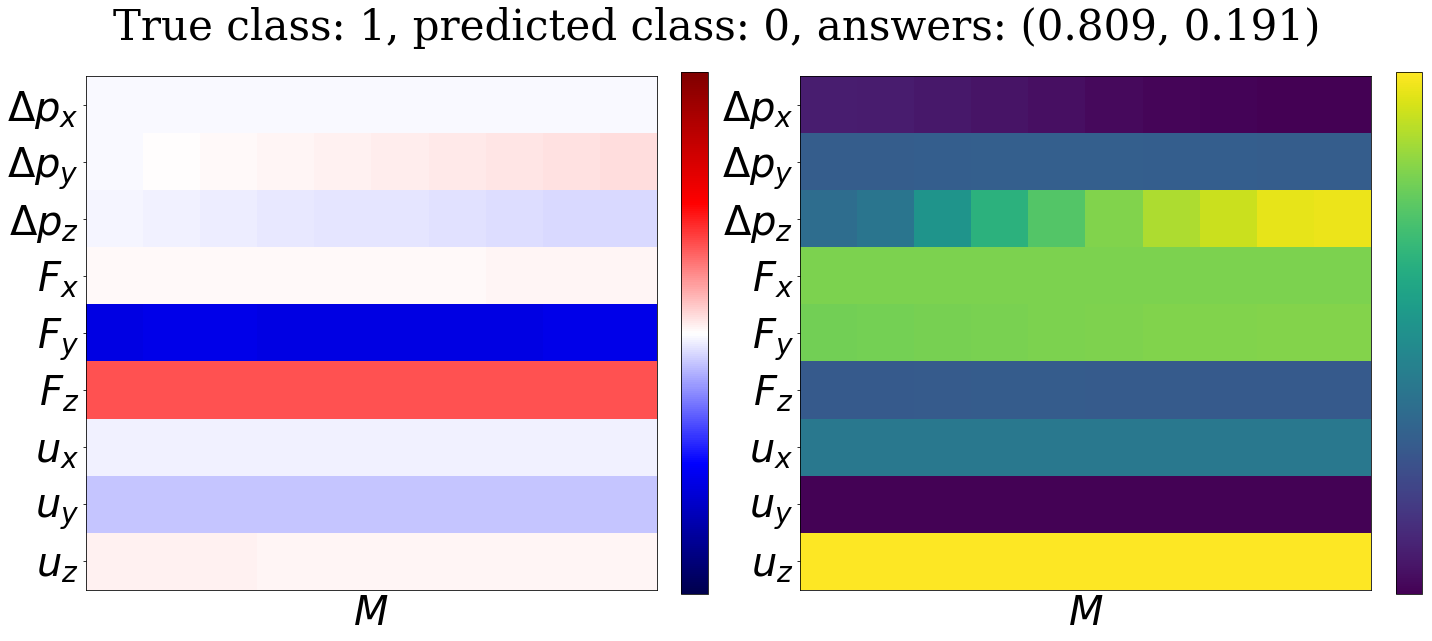}\label{fig:EventNetC1_c}}
	    \subfloat[$b+1$]{\includegraphics[width=0.25\textwidth]{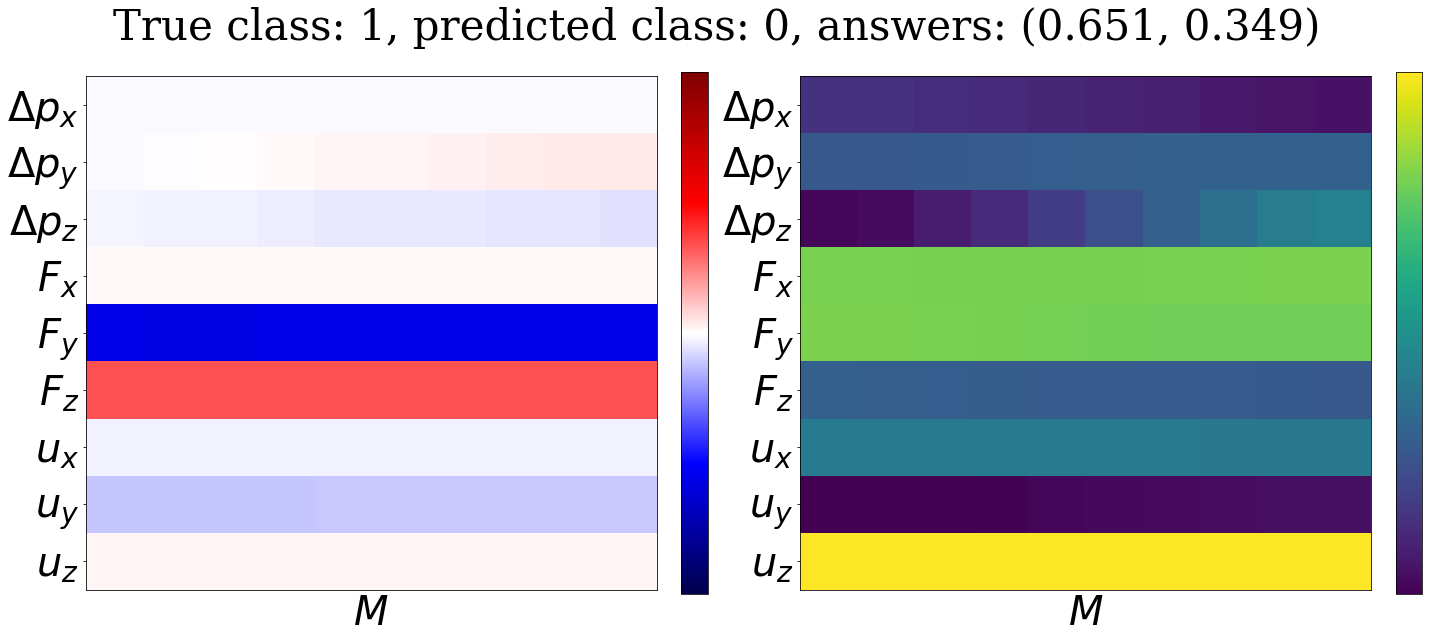}\label{fig:EventNetC1_d}}
    	
    	\subfloat[$b-4$]{\includegraphics[width=0.25\textwidth]{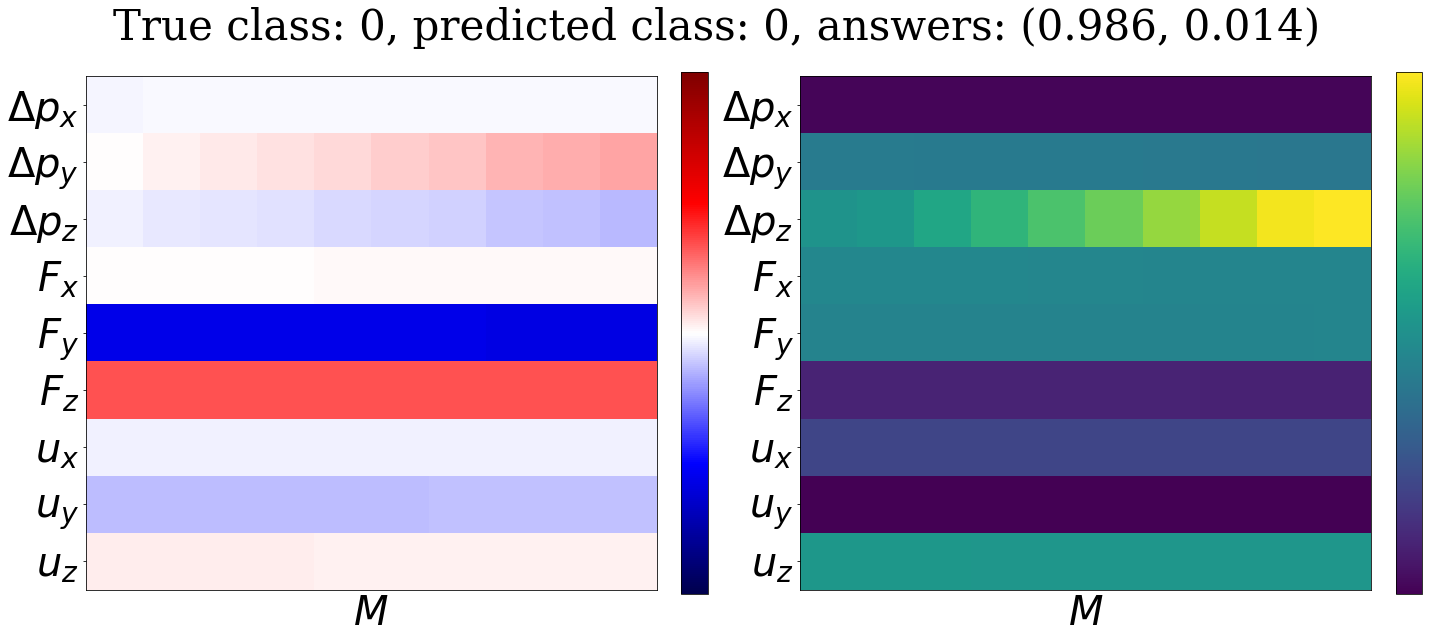}\label{fig:EventNetC2_a}}
    	\subfloat[$b-1$]{\includegraphics[width=0.25\textwidth]{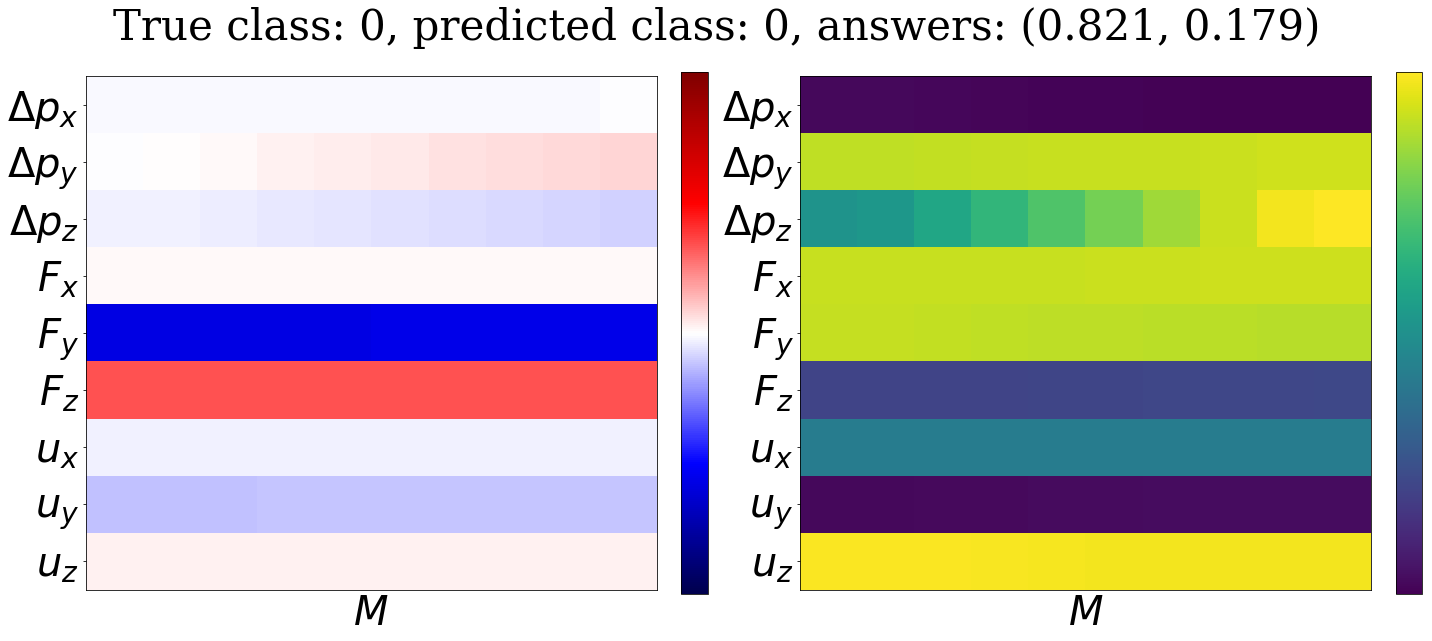}\label{fig:EventNetC2_b}}
    	\subfloat[$b$]{\includegraphics[width=0.25\textwidth]{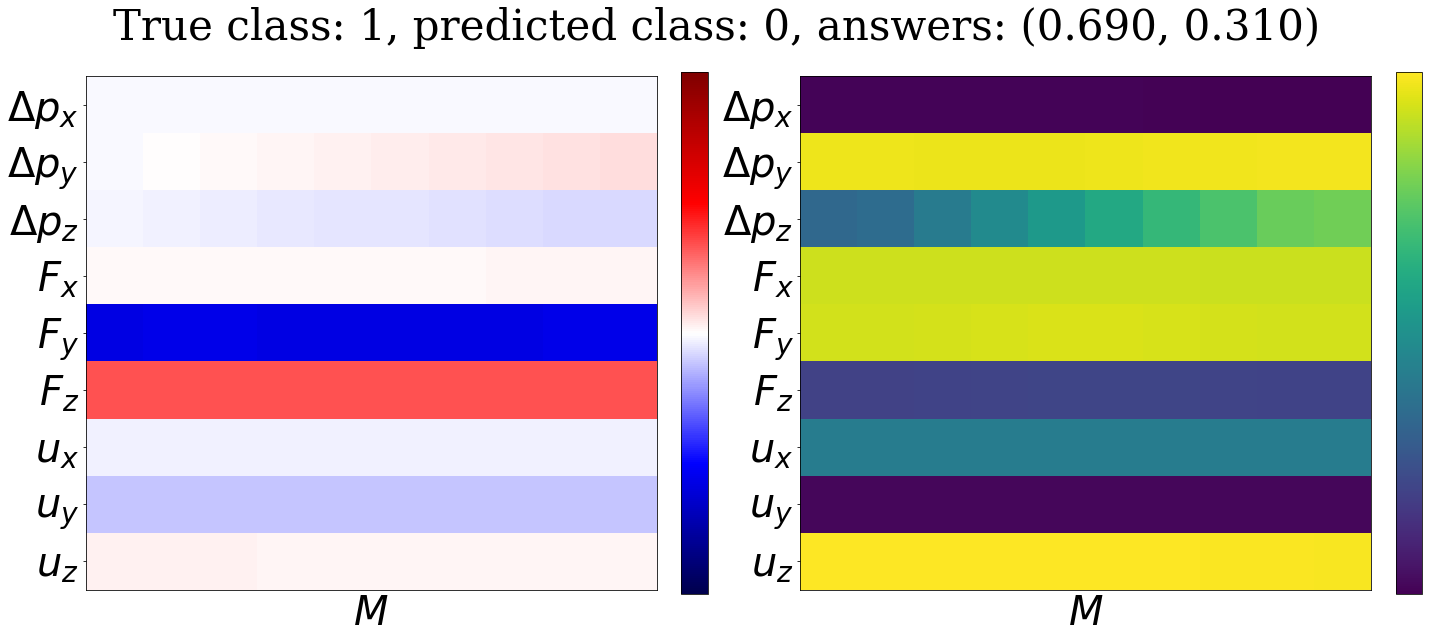}\label{fig:EventNetC2_c}}
    	\subfloat[$b+1$]{\includegraphics[width=0.25\textwidth]{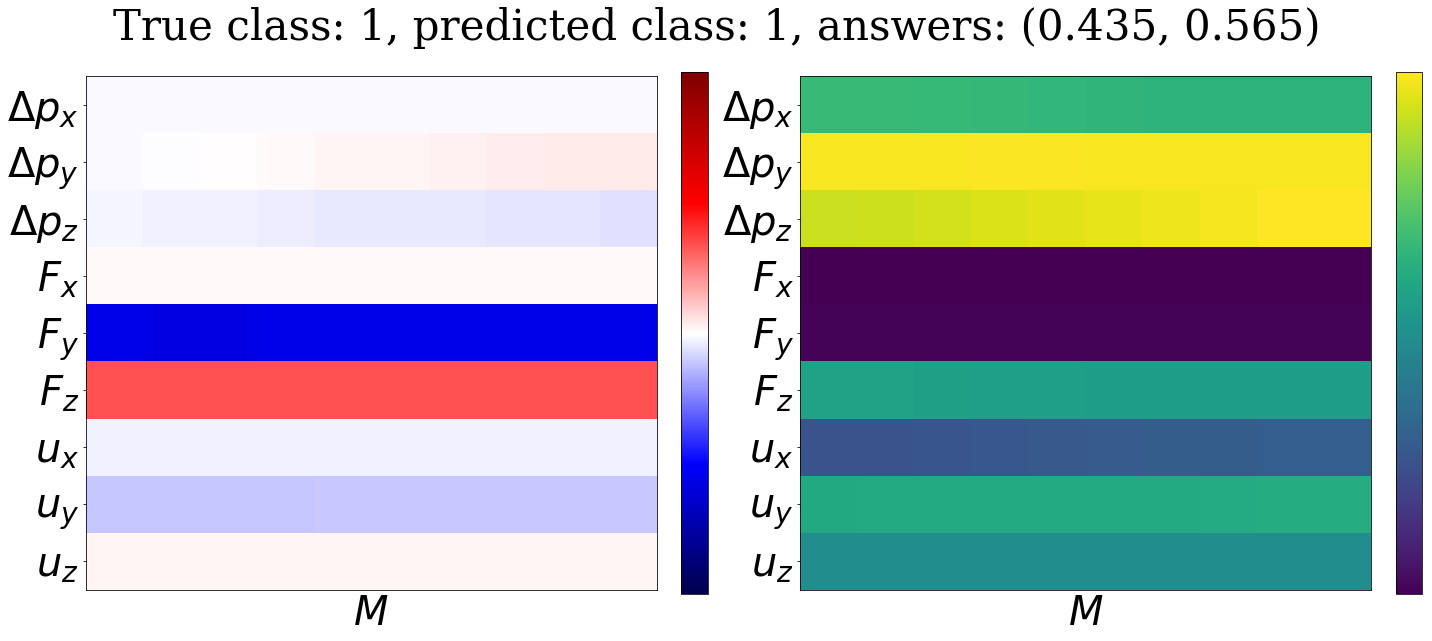}\label{fig:EventNetC2_d}}
    	
    	\subfloat[$b-4$]{\includegraphics[width=0.25\textwidth]{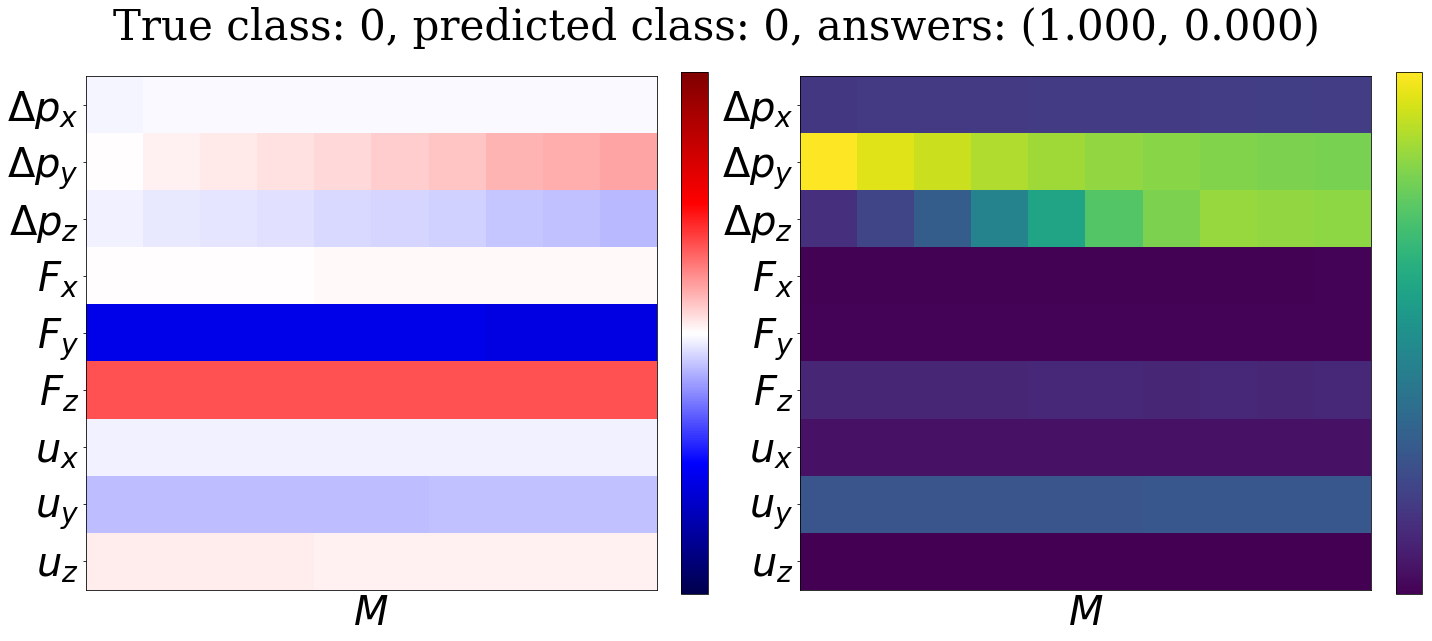}\label{fig:EventNetCLSTM_a}}
    	\subfloat[$b-1$]{\includegraphics[width=0.25\textwidth]{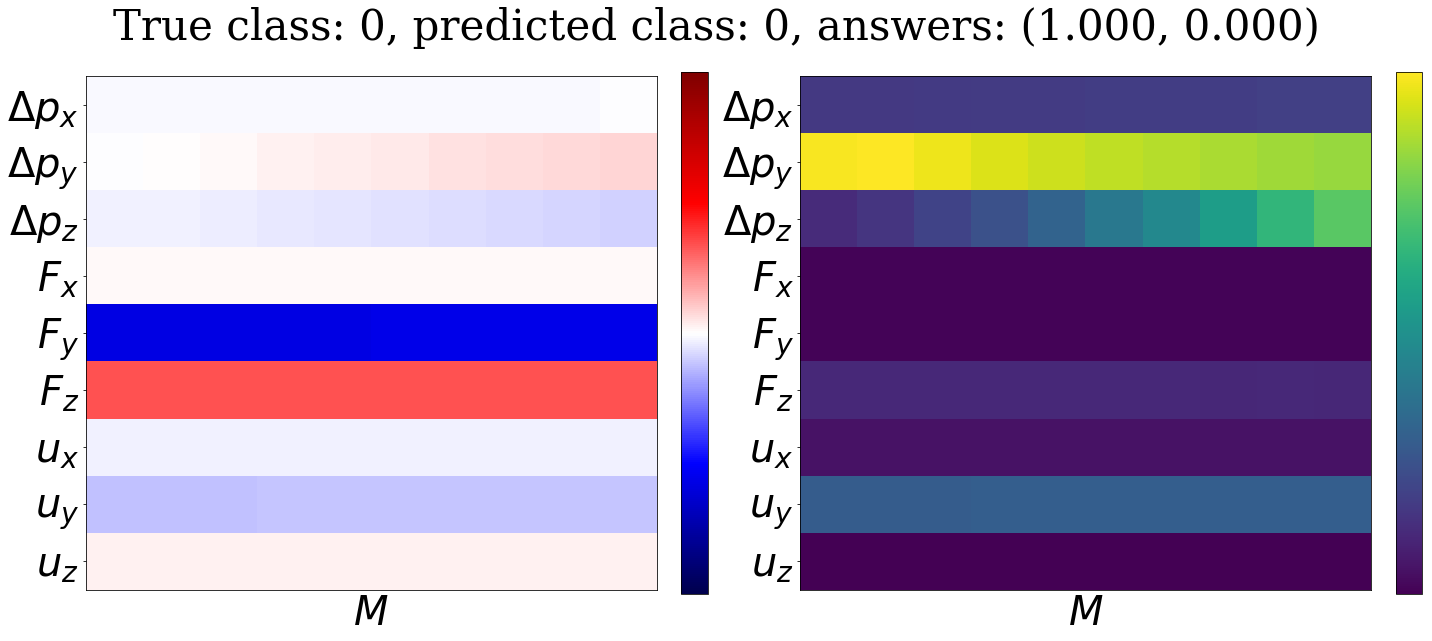}\label{fig:EventNetCLSTM_b}}
    	\subfloat[$b$]{\includegraphics[width=0.25\textwidth]{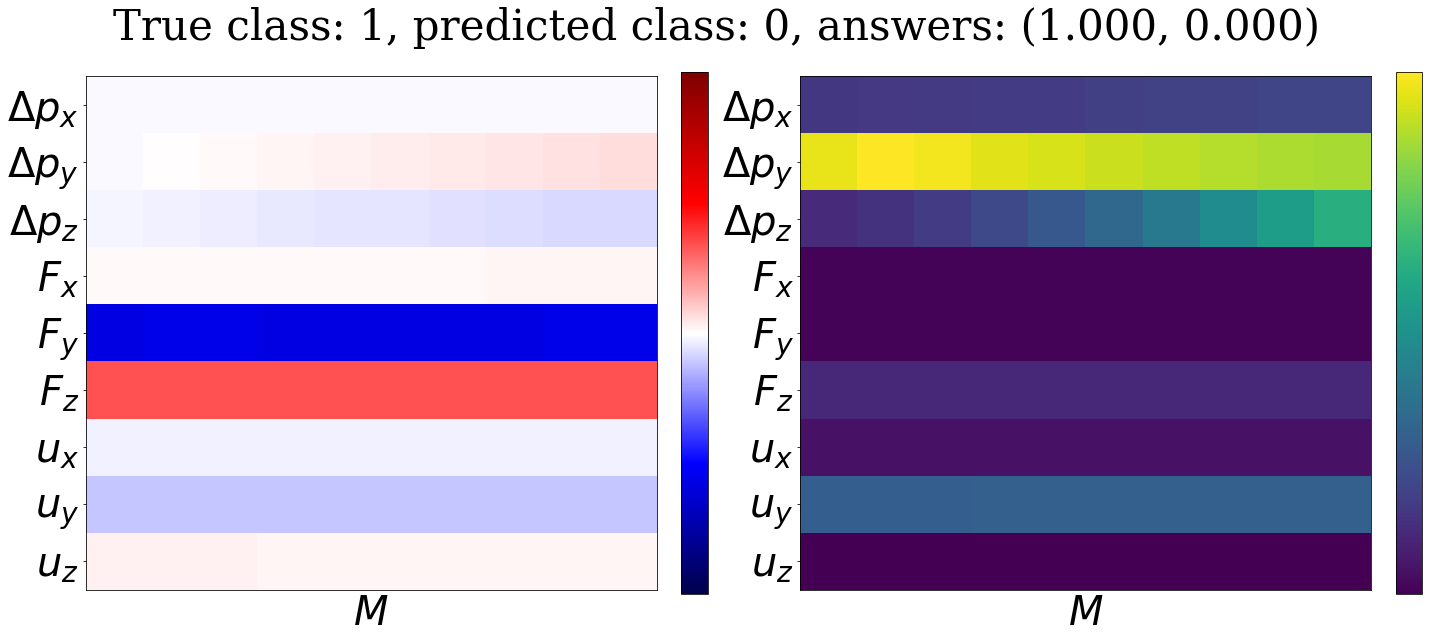}\label{fig:EventNetCLSTM_c}}
    	\subfloat[$b+1$]{\includegraphics[width=0.25\textwidth]{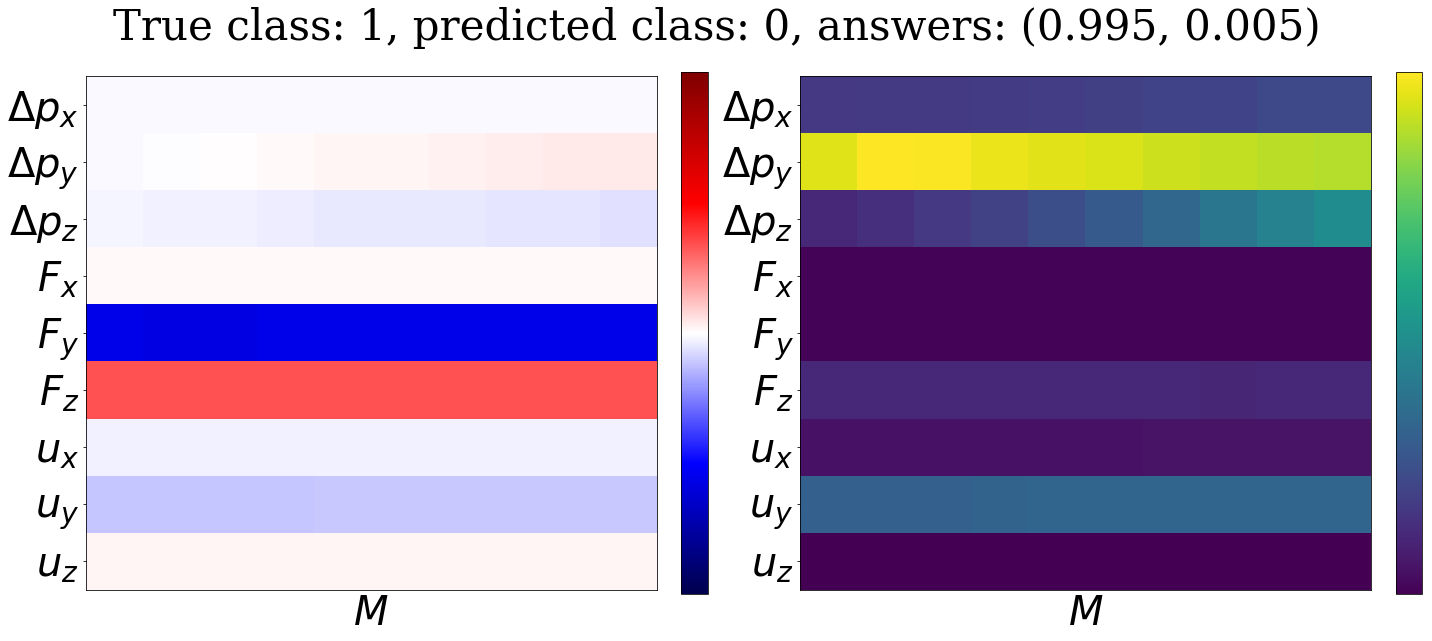}\label{fig:EventNetCLSTM_d}}
	\caption{\footnotesize Sequence of input frames at blocks $b\pm i$ during a transition for the cutting task. The actual class transitions happens during block $b$. Top row corresponds to model \texttt{EventNetC1}, the middle row to \texttt{EventNetC2}, and the bottom row to \texttt{EventNetCLSTM}. As \texttt{EventNetCLSTM} has a sequence of 3 images as input, the GradCAM images shown for it are the average of the 3 images leading up to the displayed frame. Despite their similar training performance, interpreting the models' classification decisions reveals that they all react differently to the transition.}
	\vspace{-1mm}
	\label{fig:2Dcrit} 
	\vspace{-3mm}
\end{figure*}

In the remainder of this section, we demonstrate how to utilize the proposed method to visualize and evaluate models in two examples of real-world manipulation: pushing and cutting. Pushing is a task that has been often treated with data-driven methods in the literature \cite{rodriguez,Li2018PushNetDP}. Its dynamics agree with intuition so we employ it as an instructional example. For this task, we introduce the kinodynamic images by examining the interpretations of single image frames for different classification results. Additionally, we demonstrate how to detect the features that may lead to miss-classifications during a label transition.

As opposed to pushing, which is easy to interpret but has no interaction between axes of motion, we need to also consider tasks with complicated, coupled dynamics such as cutting.
In most cases, cutting requires the knife to first break friction by sawing and then to slice downwards until the object has been cut. The coupling of the motions creates a larger degree of variation in the interaction dynamics as both are affected by the variety in object, controller and trajectory parameters. Within this task we perform three experiments.
Initially, we qualitatively show that without visualizing the decisions of trained models, the effect of small model differences on the closed-loop system can be nearly impossible to identify solely based on performance metrics. For this experiment, we compare the models \texttt{EventNetC1} and \texttt{EventNetC2} during a transition. The next two experiments are examples of quantitative, high-level evaluations of the networks' behavior. First, we compare the overall feature utilisation the models \texttt{EventNetC1} and \texttt{EventNetC2} exhibit for the entirety of the testing set. For the last experiment,  we explore the case of generalization to an unseen object and identify which features carry influence between the train and the test objects.
It should be noted that since the focus of this work is visualizing network decisions to aid in their evaluation and not to propose the best models for these tasks, we did not carry out extensive hyperparameter optimization for any of the networks. In all of the following experiments, we assume that the blocks consist of $M = 10$ timesteps. Training details as well as  training scores can be found in Tables \ref{tab:training_params} and \ref{tab:training_results}. 

\subsection{Qualitative model evaluation through frame inspection}\label{subsection:Qual}
\paragraph*{Pushing}We first consider pushing a box-like object on a plane.
Due to the geometry of the task, the kinematic and dynamic axes of interest are reflected by the $Y$ axes of the interpretable images in Fig. \ref{fig:1Dcrit}. 
To avoid making the task trivial, the network \texttt{EventNetP} is trained to detect whether the object will continue moving $H_{block} = 5$ blocks of measurement sequences in the future, countering stick-slip phenomena. The label for each input frame is determined by the relative displacement on the axis of motion during the last timestep of the block $H_{block}$ ahead. If by the end of the prediction horizon, the displacement is not larger than $0.08 \textrm{mm}$, we determined that there is no motion and assigned the negative label $C = 1$. In any other case, we considered the label positive and assigned it $C = 0$.
A transition from True Positive to True Negative can be seen in Fig. \ref{fig:1Dcrit}.


\paragraph*{Cutting}
The training goal is to predict if the cutting progress will be inhibited $H_{block}$ frames in the future and the labels are assigned similarly to the pushing task.
 The positive class ($C=0$) corresponds to normal downwards motion in $H_{block}$ blocks ahead, while the negative class ($C=1$) occurs when during the last timestep of the block of interest, the displacement $\Delta p_z$ does not surpass $0.05 \textrm{mm}$.
In this experiment, we analyze the models \texttt{EventNetC1}, \texttt{EventNetC2} and \texttt{EventNetCLSTM} during the label transition  in Fig. \ref{fig:2Dcrit}. These models have very similar scores, and for the two convolutional models the only differences are the learning rate (Table \ref{tab:training_params}) and weight initialization seed, which was random.

\subsection{Quantitative model evaluation through feature utilisation}\label{subsection:Quant}

Another important aspect of this method is that it offers a high-level overview of a model's performance. This can be easily done by calculating how many times an input feature led to a specific prediction or even by setting a threshold for the Grad-CAM activations and examining which features surpass them and contribute more.
In Fig. \ref{fig:featimport} we use the former method and compare the overall behavior of the models \texttt{EventNetC1}, \texttt{EventNetC2}, and \texttt{EventNetCLSTM} on the test set. The bar charts depict the percentage of occurrences that each feature is the main reason behind a classification result over the total cases for that category. 
Lastly, we explore the case of generalization to an unseen object by quantitatively examining the feature importance during testing on the new object.

\section{Discussion}
\subsection{Qualitative model evaluation through  frame inspection}\label{subsection:DiscQual}

In the transition example of the pushing task in Fig. \ref{fig:1Dcrit}, initially, the decision is based on the forces $F_y$ but eventually, the attention is shifted to the lack of motion $\Delta p_y$. In this sequence, the transition between classes happens at block $b$ and the correct detection at $b+4$ but not neatly as the irrelevant axis $X$ also gained importance through $\Delta p_x, F_x$. Visual inspections such as this are useful to detect models that are affected by inconsequential features and may not be robust in the presence of disturbance or noise during deployment.

Motion inhibition in cutting can be a result of inappropriate controller and trajectory parameters or the object's structure. 
The former case is the most common and usually caused by parameters that are not enough to break friction on either axis. This includes high compliance to external forces or inadequate velocity to allow progress. The latter case includes objects that structurally stop the motion because parts of them are impossible to cut through. Nonetheless, depending on the hardware limitations,
high exerted torques can also occur because of objects with dense cores, such as a carrot, or high overall density, such as a potato.

For this set of experiments, we trained the networks \texttt{EventNetC1, EventNetC2, EventNetCLSTM} to detect motion inhibition in a future block. The models have very small differences in performance as shown in Table \ref{tab:training_results}. However, visualizing and explaining their decisions shows that they respond to different patterns in the input image, reinforcing that interpretability is important to understand why two seemingly identical models can behave differently when deployed. 
Initially, in Fig. \ref{fig:EventNetC1_a}, \ref{fig:EventNetC2_a}, \ref{fig:EventNetCLSTM_a}, all networks are basing their decision to some degree on the downwards motion observed in $\Delta p_z$, with \texttt{EventNetCLSTM} placing more importance to $\Delta p_y$. One block before the transition ($b-1$) (Fig. \ref{fig:EventNetC1_b}, \ref{fig:EventNetC2_b}, \ref{fig:EventNetCLSTM_b}), only \texttt{EventNetC2} shifts its attention to the lack of motion on $\Delta p_y$ and the corresponding force $F_y$ which indicates the blade is not able to saw through the object, while \texttt{EventNetCLSTM} continues to only focus on the displacement.
At the time of the actual transition (Fig. \ref{fig:EventNetC1_c}, \ref{fig:EventNetC2_c}, \ref{fig:EventNetCLSTM_c}), no model detects it immediately. \texttt{EventNetCLSTM} does not change its behaviour and \texttt{EventNetC2} is considering the less important lateral forces $F_x$ equally as much as $F_y$ while \texttt{EventNetC1} is focusing on axis $Z$ through $\Delta p_z$ and $u_z$. Finally, at the immediate next block (Fig. \ref{fig:EventNetC1_d}, \ref{fig:EventNetC2_d}, \ref{fig:EventNetCLSTM_d}), \texttt{EventNetC1} falsely disregards $\Delta p_z$ and miss-classifies the input while \texttt{EventNetC2} focuses on the lack of displacement and correctly classifies it. \texttt{EventNetCLSTM} also miss-classifies the input, basing its decision mostly on the lack of displacement. It is not until block $b+3$ that it considers the forces and correctly classify the sample, potentially because the network is operating recurrently on a sequence of the 3 most recent frames. This does not offer the model any more information closer to the prediction horizon than its CNN counterparts, but does encourage it to focus on trends across longer sequences. In the integrated system, this difference in reasoning could be reflected in delayed response time or sensitivity to non crucial features which would in turn negatively affect any module determining the corrective action and the overall task success.

\subsection{Quantitative model evaluation through feature utilisation}\label{subsection:DiscQuant}
\begin{figure}
	\centering
    	\subfloat[\footnotesize \texttt{EventNetC1}]{\includegraphics[width=0.95\linewidth]{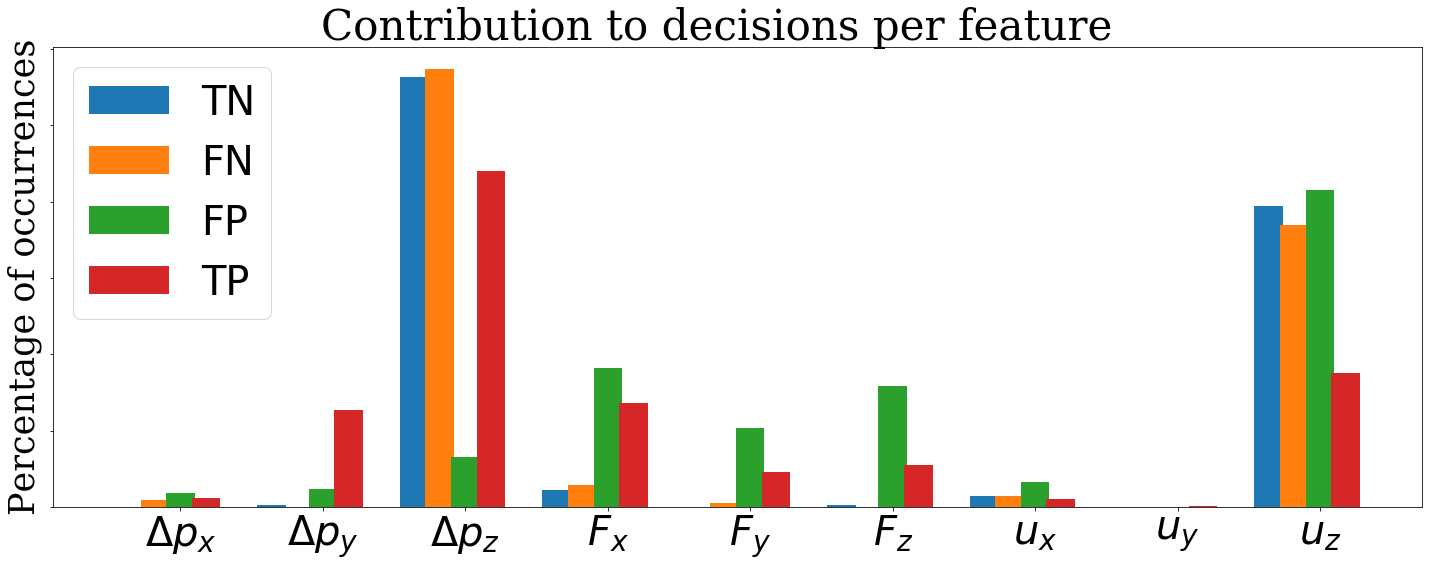}\label{fig:featimport_net1}} \vspace{1pt} \\
    	\subfloat[\footnotesize \texttt{EventNetC2}]{\includegraphics[width=0.95\linewidth]{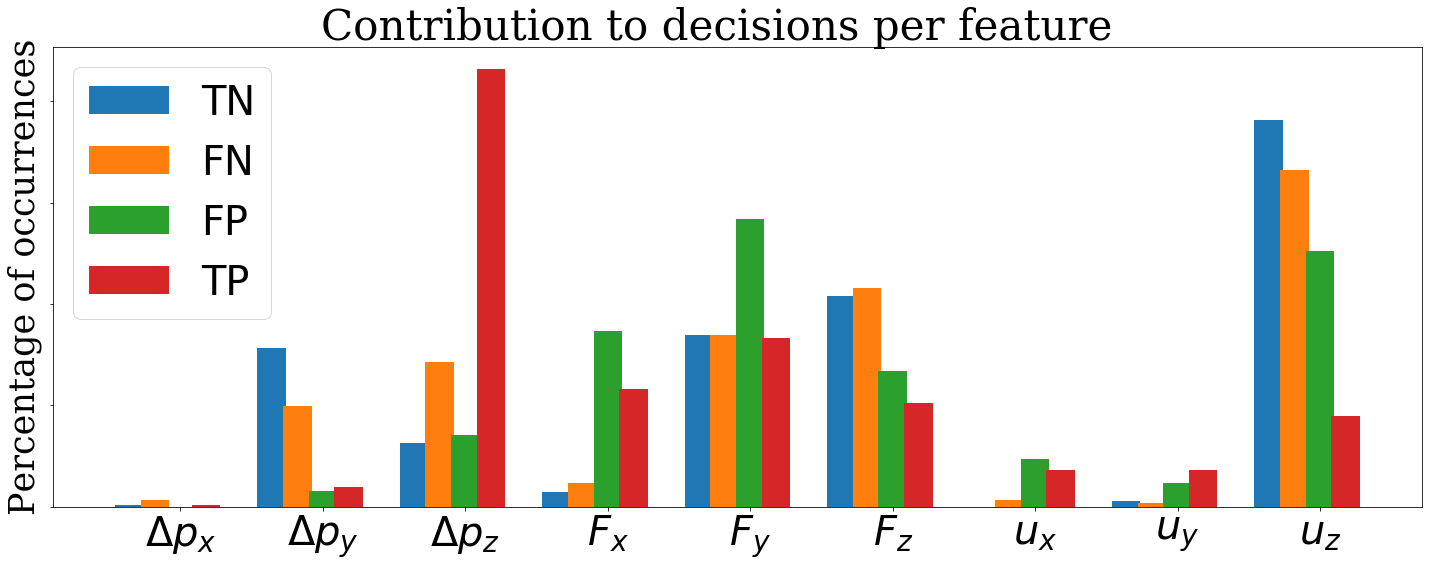}\label{fig:featimport_net2}} \vspace{1pt} \\
    	\subfloat[\footnotesize
    	\texttt{EventNetCLSTM}]{\includegraphics[width=0.95\linewidth]{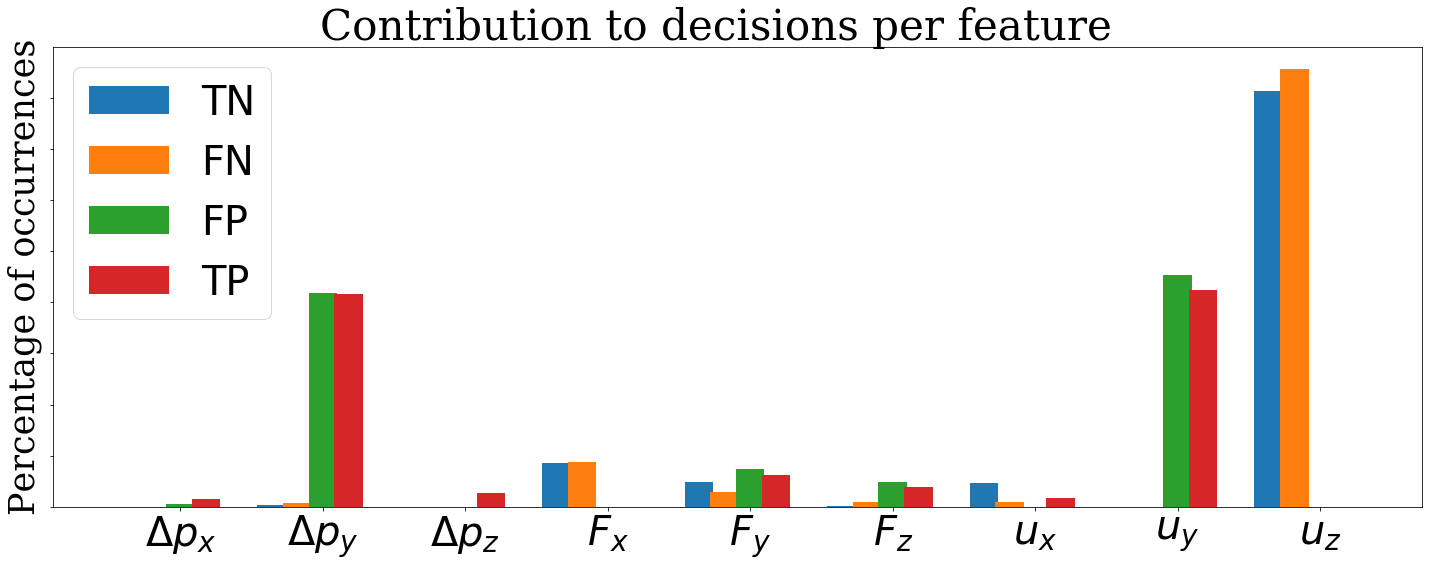}\label{fig:featimport_CLSTMNet}} 
	\caption{\footnotesize A comparison over features prioritized per classification result. Each bar represents the percentage of feature instances over the total cases of every category. Total occurrences TP:1166, TN:1235, FN:214, FP:211.}
	\label{fig:featimport} 
	\vspace{-5mm}
\end{figure}

Fig. \ref{fig:featimport} depicts the overall feature utilizations for the entirety of the testing set and concisely presents the different patterns favored by the models. 

For \texttt{EventNetC1} it is clear from Fig. \ref{fig:featimport_net1} that the most used feature is the downwards displacement $\Delta p_z$ both for True Negative (TN) and True Positive (TP) predictions. The control input on the same axis is the second most common indication for TP while the sawing axis is only slightly used, indicating a one-sided view of the task. On the other hand, \texttt{EventNetC2} decidedly utilizes $\Delta p_z$ for TP. The obstructed motion (TN) is mostly detected through $u_z$ along with influence from forces $F_y, F_z$ and displacement $\Delta p_y$ which better reflects the nature of the task.
As also exhibited in \ref{fig:2Dcrit}, \texttt{EventNetCLSTM} bases its predictions of unhindered movement (TP) mostly on $\Delta p_y, u_y$, and bases TN/FN samples almost solely on the z axis control input.

  \begin{figure}
    \includegraphics[width = 0.95\linewidth]{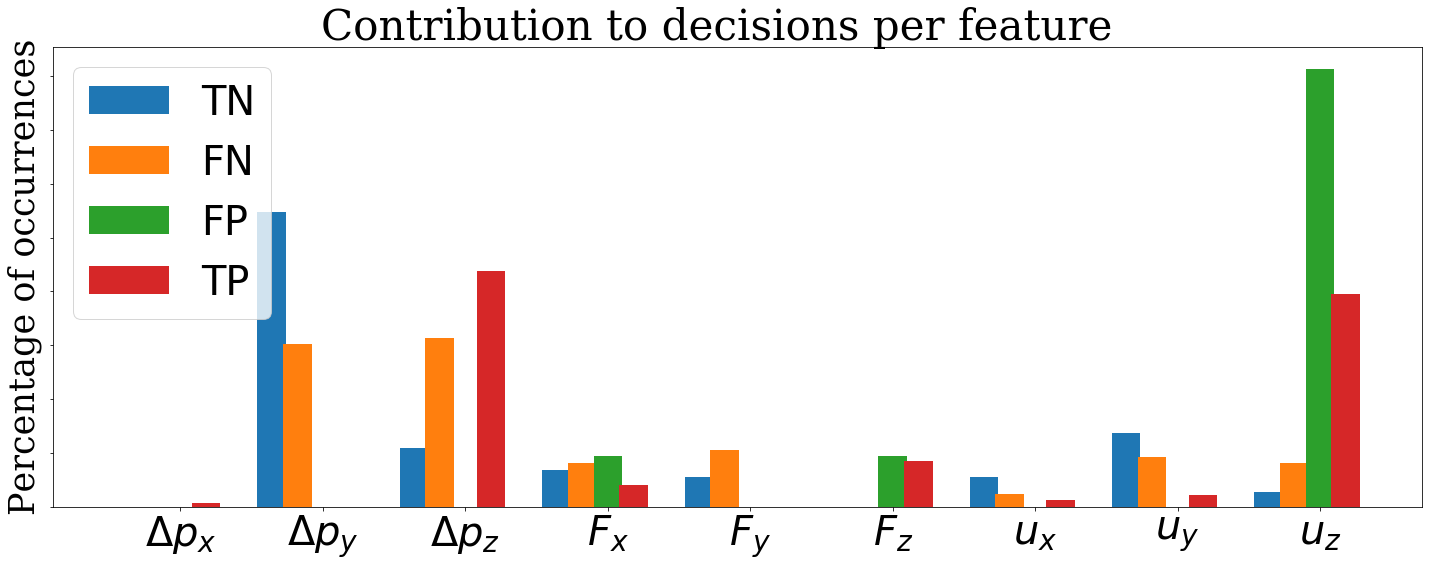}
    \caption{\footnotesize Features prioritized per classification result for generalization to unseen object. Each bar represents the percentage of feature instances over the total cases of every category. Total occurrences TP:73, TN:725, FN:32, FP:86.}
        \label{fig:gen_results}
        \vspace{-7mm}
\end{figure}

Lastly, we evaluated which features are important for generalization to an object with distinct dynamics.
For this experiment, we excluded one object category from the training set (zucchinis), trained \texttt{EventNetGen} on the remainder (potatoes, carrots) and then tested on the unseen object. Given the differences in texture and consistency, it is not surprising that the model has difficulty detecting when the motion ceases as demonstrated by the training results in Table \ref{tab:training_results}. It is interesting to notice however, that the different motion profiles are the main reasons behind FN but the control input $u_z$ effectively dominates the FP results.

 This is due to the fact that the testing object is significantly easier to cut. Since it would not cause a mechanical shutdown, during data collection we were commanding more aggressive trajectories, which in turn led to higher control inputs. 
 In the case of the training set objects, these inputs would probably mean that the knife encountered too much resistance and was not able to move, which justifies the missclassification and lack of generalization performance.


\section{Conclusion}
In this work, we presented a framework that instills interpretability in manipulation tasks with kinematic and dynamic data. We achieved that by transcribing sequences of data in visually intuitive images that encode the temporal evolution of the features. Through our approach, it is possible to explain the decisions of learned models used in any robotic task that is formulated as a classification problem and utilizes synchronous features. Without loss of generality, we treated translational motion and forces and binary classification. However, tasks with different dimensions or types of features can be explained simply by modifying what comprises the height of the image, $L$. In addition, multi-label classification can be treated exactly in the same manner as Grad-CAM is not constrained by the number of classes.
In our experiments we showed how to detect features that lead to miss-classifications by inspecting isolated sequences during label transitions. Furthermore, we showed how interpretability can help distinguish models that are seemingly identical based on their learning scores but will behave differently during deployment. Lastly, we illustrated how the same pipeline can be used to derive quantitative results about feature utilization and how that can be applied to examine the model's generalization abilities.
However, this method is versatile and can be used for more tasks. For example, for analyzing if architectures with different complexities are in essence deciding based on the same features or to monitor if the changes made in training, consistently produce models with similar performance and feature utilization.
Note that the colormap choice can affect the training results. For a task like pushing, all the quantities have the same sign and sequential colormaps are appropriate. However, in other cases the features take both positive and negative values that lead to different dynamics and diverging colormaps are preferable. During training, we observed that the neutral zone color could also bias the network and lead to different results.
Currently, this method is only applicable to classification tasks. In the future, we aim to explore more interpretability modules and extend it to continuous outputs which will encompass regression tasks and learned dynamics models. 


\bibliographystyle{IEEEtran}
\bibliography{example}

\begin{thebibliography}{10}
\providecommand{\url}[1]{#1}
\csname url@samestyle\endcsname
\providecommand{\newblock}{\relax}
\providecommand{\bibinfo}[2]{#2}
\providecommand{\BIBentrySTDinterwordspacing}{\spaceskip=0pt\relax}
\providecommand{\BIBentryALTinterwordstretchfactor}{4}
\providecommand{\BIBentryALTinterwordspacing}{\spaceskip=\fontdimen2\font plus
\BIBentryALTinterwordstretchfactor\fontdimen3\font minus
  \fontdimen4\font\relax}
\providecommand{\BIBforeignlanguage}[2]{{%
\expandafter\ifx\csname l@#1\endcsname\relax
\typeout{** WARNING: IEEEtran.bst: No hyphenation pattern has been}%
\typeout{** loaded for the language `#1'. Using the pattern for}%
\typeout{** the default language instead.}%
\else
\language=\csname l@#1\endcsname
\fi
#2}}
\providecommand{\BIBdecl}{\relax}
\BIBdecl

\bibitem{code}
I.~Mitsioni and J.~Mänttäri, ``Intepretability in contact-rich manipulation
  via kinodynamic images,''
  \emph{\url{https://github.com/imitsioni/interpretable_manipulation}}.

\bibitem{bohgGraspSurvey}
J.~{Bohg}, A.~{Morales}, T.~{Asfour}, and D.~{Kragic}, ``Data-driven grasp
  synthesis—a survey,'' \emph{IEEE Transactions on Robotics}, vol.~30, no.~2,
  pp. 289--309, 2014.

\bibitem{slipDet}
I.~{Agriomallos}, S.~{Doltsinis}, I.~{Mitsioni}, and Z.~{Doulgeri}, ``Slippage
  detection generalizing to grasping of unknown objects using machine learning
  with novel features,'' \emph{IEEE Robotics and Automation Letters}, vol.~3,
  no.~2, pp. 942--948, 2018.

\bibitem{PomdpHapticData}
J.~{Sung}, J.~K. {Salisbury}, and A.~{Saxena}, ``Learning to represent haptic
  feedback for partially-observable tasks,'' in \emph{2017 IEEE International
  Conference on Robotics and Automation (ICRA)}, 2017, pp. 2802--2809.

\bibitem{dressing}
Z.~{Erickson}, H.~M. {Clever}, G.~{Turk}, C.~K. {Liu}, and C.~C. {Kemp}, ``Deep
  haptic model predictive control for robot-assisted dressing,'' in \emph{2018
  IEEE International Conference on Robotics and Automation (ICRA)}, 2018, pp.
  4437--4444.

\bibitem{Nagabandi2019}
A.~Nagabandi, K.~Konoglie, S.~Levine, and V.~Kumar, ``{Deep Dynamics Models for
  Learning Dexterous Manipulation},'' \emph{Conference on Robot Learning
  (CoRL)}, pp. 1--12, 2019.

\bibitem{openAIinhand}
O.~M. Andrychowicz, B.~Baker, M.~Chociej, R.~Józefowicz, B.~McGrew,
  J.~Pachocki, A.~Petron, M.~Plappert, G.~Powell, A.~Ray, J.~Schneider,
  S.~Sidor, J.~Tobin, P.~Welinder, L.~Weng, and W.~Zaremba, ``Learning
  dexterous in-hand manipulation,'' \emph{The International Journal of Robotics
  Research}, vol.~39, no.~1, pp. 3--20, 2020.

\bibitem{GradCam}
R.~R. Selvaraju, M.~Cogswell, A.~Das, R.~Vedantam, D.~Parikh, and D.~Batra,
  ``Grad-cam: Visual explanations from deep networks via gradient-based
  localization,'' \emph{International Journal of Computer Vision}, vol. 128,
  no.~2, p. 336–359, Oct 2019.

\bibitem{learningTactile1}
B.~{Sundaralingam}, A.~S. {Lambert}, A.~{Handa}, B.~{Boots}, T.~{Hermans},
  S.~{Birchfield}, N.~{Ratliff}, and D.~{Fox}, ``Robust learning of tactile
  force estimation through robot interaction,'' in \emph{2019 International
  Conference on Robotics and Automation (ICRA)}, 2019, pp. 9035--9042.

\bibitem{learningTactile2}
N.~F. {Lepora}, A.~{Church}, C.~{de Kerckhove}, R.~{Hadsell}, and J.~{Lloyd},
  ``From pixels to percepts: Highly robust edge perception and contour
  following using deep learning and an optical biomimetic tactile sensor,''
  \emph{IEEE Robotics and Automation Letters}, vol.~4, no.~2, pp. 2101--2107,
  2019.

\bibitem{gelsight}
S.~{Tian}, F.~{Ebert}, D.~{Jayaraman}, M.~{Mudigonda}, C.~{Finn},
  R.~{Calandra}, and S.~{Levine}, ``Manipulation by feel: Touch-based control
  with deep predictive models,'' in \emph{2019 International Conference on
  Robotics and Automation (ICRA)}, 2019, pp. 818--824.

\bibitem{makingsense}
M.~A. {Lee}, Y.~{Zhu}, K.~{Srinivasan}, P.~{Shah}, S.~{Savarese}, L.~{Fei-Fei},
  A.~{Garg}, and J.~{Bohg}, ``Making sense of vision and touch: Self-supervised
  learning of multimodal representations for contact-rich tasks,'' in
  \emph{2019 International Conference on Robotics and Automation (ICRA)}, 2019,
  pp. 8943--8950.

\bibitem{DeepInsight}
A.~Sharma, E.~Vans, D.~Shigemizu, K.~A. Boroevich, and T.~Tsunoda,
  ``Deepinsight: A methodology to transform a non-image data to an image for
  convolution neural network architecture,'' \emph{Nature Scientific Reports},
  vol. 11399, no.~9, Aug 2019.

\bibitem{Simonyan2014DeepMaps}
K.~Simonyan, A.~Vedaldi, and A.~Zisserman, ``Deep inside convolutional
  networks: Visualising image classification models and saliency maps,'' in
  \emph{Workshop at International Conference on Learning Representations},
  2014.

\bibitem{Zeiler2014}
M.~D. Zeiler and R.~Fergus, ``{Visualizing and understanding convolutional
  networks},'' \emph{Lecture Notes in Computer Science (including subseries
  Lecture Notes in Artificial Intelligence and Lecture Notes in
  Bioinformatics)}, vol. 8689 LNCS, no. PART 1, pp. 818--833, 2013.

\bibitem{Montavon2018}
G.~Montavon, W.~Samek, and K.~R. M{\"{u}}ller, ``{Methods for interpreting and
  understanding deep neural networks},'' \emph{Digital Signal Processing: A
  Review Journal}, vol.~73, pp. 1--15, 2018.

\bibitem{tcavglobal_icml18}
B.~{Kim}, W.~M., J.~{Gilmer}, C.~C., W.~J., , F.~{Viegas}, and R.~{Sayres}, ``{
  Interpretability Beyond Feature Attribution: Quantitative Testing with
  Concept Activation Vectors (TCAV) },'' \emph{ICML}, 2018.

\bibitem{Erhan2009VisualizingNetwork}
D.~Erhan, Y.~Bengio, A.~Courville, and P.~Vincent, ``{Visualizing higher-layer
  features of a deep network},'' \emph{Bernoulli}, no. 1341, pp. 1--13, 2009.

\bibitem{ExcitationBackprop16}
\BIBentryALTinterwordspacing
J.~Zhang, Z.~Lin, J.~Brandt, X.~Shen, and S.~Sclaroff, ``Top-down neural
  attention by excitation backprop,'' \emph{CoRR}, vol. abs/1608.00507, 2016.
  [Online]. Available: \url{http://arxiv.org/abs/1608.00507}
\BIBentrySTDinterwordspacing

\bibitem{adebayoneurips18}
\BIBentryALTinterwordspacing
J.~Adebayo, J.~Gilmer, M.~Muelly, I.~J. Goodfellow, M.~Hardt, and B.~Kim,
  ``Sanity checks for saliency maps,'' \emph{CoRR}, vol. abs/1810.03292, 2018.
  [Online]. Available: \url{http://arxiv.org/abs/1810.03292}
\BIBentrySTDinterwordspacing

\bibitem{hristov2019disentangled}
Y.~Hristov, D.~Angelov, M.~Burke, A.~Lascarides, and S.~Ramamoorthy,
  ``Disentangled relational representations for explaining and learning from
  demonstration,'' in \emph{Conference on Robot Learning (CoRL)}, 2019.

\bibitem{kulkarni2020designing}
A.~Kulkarni, S.~Sreedharan, S.~Keren, T.~Chakraborti, D.~Smith, and
  S.~Kambhampati, ``Designing environments conducive to interpretable robot
  behavior,'' in \emph{2020 IEEE/RSJ International Conference on Intelligent
  Robots and Systems (IROS)}, 2020.

\bibitem{interpretableCollab}
B.~{Hayes} and J.~A. {Shah}, ``Interpretable models for fast activity
  recognition and anomaly explanation during collaborative robotics tasks,'' in
  \emph{2017 IEEE International Conference on Robotics and Automation (ICRA)},
  2017, pp. 6586--6593.

\bibitem{dotToDot}
B.~{Beyret}, A.~{Shafti}, and A.~A. {Faisal}, ``Dot-to-dot: Explainable
  hierarchical reinforcement learning for robotic manipulation,'' in \emph{2019
  IEEE/RSJ International Conference on Intelligent Robots and Systems (IROS)},
  2019, pp. 5014--5019.

\bibitem{prospection}
C.~{Paxton}, Y.~{Bisk}, J.~{Thomason}, A.~{Byravan}, and D.~{Foxl},
  ``Prospection: Interpretable plans from language by predicting the future,''
  in \emph{2019 International Conference on Robotics and Automation (ICRA)},
  2019, pp. 6942--6948.

\bibitem{Ehsani_2020_CVPR}
K.~Ehsani, S.~Tulsiani, S.~Gupta, A.~Farhadi, and A.~Gupta, ``Use the force,
  luke! learning to predict physical forces by simulating effects,'' in
  \emph{Proceedings of the IEEE/CVF Conference on Computer Vision and Pattern
  Recognition (CVPR)}, June 2020.

\bibitem{wormdude}
M.~{Lechner}, R.~{Hasani}, M.~{Zimmer}, T.~A. {Henzinger}, and R.~{Grosu},
  ``Designing worm-inspired neural networks for interpretable robotic
  control,'' in \emph{2019 International Conference on Robotics and Automation
  (ICRA)}, 2019, pp. 87--94.

\bibitem{mitsioni2020modelling}
\BIBentryALTinterwordspacing
I.~Mitsioni, Y.~Karayiannidis, and D.~Kragic, ``Modelling and learning dynamics
  for robotic food-cutting,'' vol. abs/2003.09179, 2020. [Online]. Available:
  \url{https://arxiv.org/abs/2003.09179}
\BIBentrySTDinterwordspacing

\bibitem{rodriguez}
M.~{Bauza} and A.~{Rodriguez}, ``A probabilistic data-driven model for planar
  pushing,'' in \emph{2017 IEEE International Conference on Robotics and
  Automation (ICRA)}, 2017, pp. 3008--3015.

\bibitem{Li2018PushNetDP}
J.~Li, W.~S. Lee, and D.~Hsu, ``Push-net: Deep planar pushing for objects with
  unknown physical properties,'' in \emph{Robotics: Science and Systems}, 2018.

\end{thebibliography}

\end{document}